%% file: paper.tex
\documentclass{article}

\PassOptionsToPackage{numbers,sort&compress}{natbib}
\usepackage{dualverse}

\usepackage[utf8]{inputenc}
\usepackage[T1]{fontenc}
\usepackage{lmodern}

\usepackage{hyperref}
\usepackage{url}
\usepackage{booktabs}
\usepackage{array}
\usepackage{amsfonts}
\usepackage{amsmath}
\usepackage{amssymb}
\usepackage{amsthm}
\usepackage{nicefrac}
\usepackage{microtype}
\usepackage{xcolor}
\usepackage{graphicx}
\usepackage{placeins}
\usepackage{multirow}
\usepackage[capitalize,noabbrev]{cleveref}

\title{Low-Bit N-Gram Embeddings for Language Models}

\SetPaperDate{22-Oct-2026}
\SetCorrespondence{info@dualverse.ai}
\SetPaperLogo{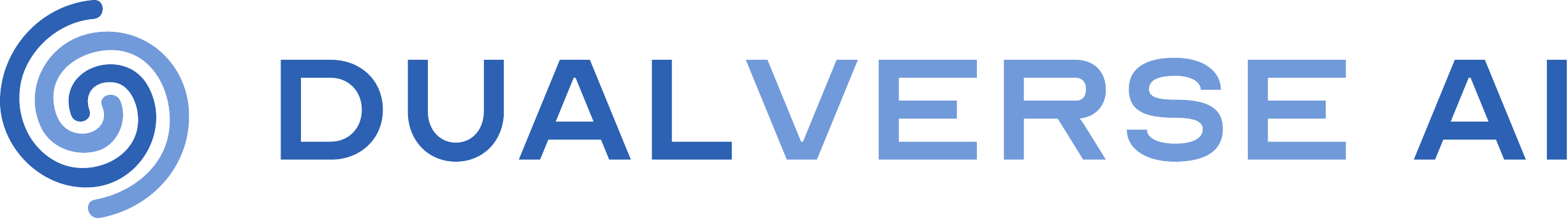}
\SetPDFAuthor{Stephen Chung, Wenyu Du}
\SetHeaderTitle{Low-Bit N-Gram Embeddings for Language Models}
\SetDualverseAuthors{%
  Stephen Chung\\DualverseAI%
  \and Wenyu Du\\DualverseAI%
}{%
  {\sffamily\bfseries\color{DualverseA}%
    Stephen Chung \quad\quad Wenyu Du%
  }\\[1em]
  \parbox{0.9\linewidth}{\centering
    \normalfont\sffamily\bfseries\small\color{DualverseB}%
    DualverseAI%
  }%
}

\begin{document}

\input{sections/abstract}
\maketitle
\begingroup
\renewcommand{\thefootnote}{}
\footnotetext[1]{This work extends a human-verified finding made by an rStation agent (\mbox{\url{https://rstation.ai/}}) as part of our \href{https://dualverse.ai/auto_research}{Autonomous Research Program} (\mbox{\url{https://dualverse.ai/auto_research}}). See Section~\ref{sec:ai_use} for a discussion of AI use.}
\endgroup

\vspace{1.8ex}
\input{sections/introduction}
\input{sections/method}
\clearpage
\input{sections/results}
\input{sections/related}
\input{sections/conclusion}
\input{sections/ai_use}

\clearpage
\bibliographystyle{unsrtnat}
\bibliography{references}

\newpage
\appendix
\renewcommand{\theHtable}{A\arabic{table}}
\renewcommand{\theHfigure}{A\arabic{figure}}
\input{sections/appendix}

\end{document}

%% file: sections/abstract.tex
\begin{abstract}
Many recent open-weight language models use n-gram embedding tables to improve performance. These tables store learned representations of short token sequences, such as pairs or triples of consecutive tokens, which the model retrieves during processing. However, they can require substantial memory: DeepSeek-V4.1-Flash, for example, allocates 196 billion parameters to its Engram n-gram memory. We therefore investigate whether this memory cost can be reduced while preserving the performance improvement that the tables provide. We propose storing n-gram embeddings at low precision and systematically investigate different quantization methods and bit allocations across tables. Across nanochat-, Llama-, and GPT-2-family models trained from scratch, four-bit tables save approximately 75\% of table memory while preserving more than 99\% of the validation-loss improvement from adding the tables. Using fewer bits leads to a larger drop in performance, although we find that a short fine-tuning phase recovers a significant part of this loss. In addition, reinvesting the saved memory in larger tables can improve performance beyond that of the original full-precision tables while still using less memory. These table memory savings also extend to large existing models: on DeepSeek-V4.1-Flash and Qwen3.8-Flash-Next, low-bit tables save 47–55\% of table memory, with perplexity increases of less than 1.2\% and largely preserved downstream task performance. These findings establish low-bit precision as a practical way to reduce the memory requirements of n-gram embeddings and use the saved capacity to improve model performance.
\end{abstract}

%% file: sections/introduction.tex
\section{Introduction}
\label{sec:intro}

Recent studies have shown that n-gram embedding tables can improve language model performance with little additional computation~\citep{huang2025overtokenized,cheng2026dse}. These tables store learned representations of short token sequences, such as pairs or triples of consecutive tokens. At each position, the model retrieves the corresponding representations to supplement its processing of the text. Because each lookup accesses only a few embeddings, the tables can add substantial model capacity while keeping the computation per token small. This approach has been adopted in DeepSeek's Engram memory, which contains 196 billion parameters in DeepSeek-V4.1-Flash~\citep{deepseek2026v41flash}. However, storing these large tables requires substantial memory, motivating methods that preserve their performance benefits with a smaller memory footprint.

The memory cost arises from the enormous number of possible token combinations. To accommodate them, existing methods map n-grams into a fixed number of table embeddings, allowing multiple n-grams to share the same embedding~\citep{huang2025overtokenized,cheng2026dse}. Smaller tables force more unrelated sequences to share representations, which can reduce their benefit to the model. Increasing the number of embeddings alleviates this problem but requires more memory. We therefore investigate a different way to reduce table memory: keeping the embeddings while using fewer bits for each embedding value. This also raises an opportunity: if lower precision preserves most of the tables' benefit, the saved memory could support more embeddings and ultimately improve performance.

We propose compressing n-gram embedding tables by storing each value at low precision. At such low precision, the choice of representable values can affect how well the embeddings are preserved. We therefore systematically compare evenly spaced values, values chosen from normal quantiles, and values fitted to the learned embedding distribution using Lloyd--Max quantization~\citep{lloyd1982least,max1960quantizing}. We also examine whether bigram and trigram tables benefit from different precision levels, allowing us to allocate bits according to their sensitivity to compression. Together, these experiments offer practical guidance on choosing quantization methods and bit allocations to balance table memory and model performance.

We evaluate this approach on nanochat-, Llama-, and GPT-2-family models trained from scratch~\citep{karpathy2025nanochat,karpathy2026autoresearch,touvron2023llama,radford2019gpt2}. We train each model with and without n-gram tables to establish how much the tables improve performance. We then compress the trained tables to different precisions and measure the table memory saved and the improvement retained. Beyond compression alone, we investigate whether a short fine-tuning phase can recover lost performance and whether reinvesting the saved memory in additional embeddings can outperform the original full-precision tables. To test whether these savings extend to large existing models, we also compress the tables in DeepSeek-V4.1-Flash and Qwen3.8-Flash-Next, saving 47--55\% of table memory while largely preserving performance on downstream tasks.

\begin{figure}[t]
  \centering
  \includegraphics[width=\linewidth]{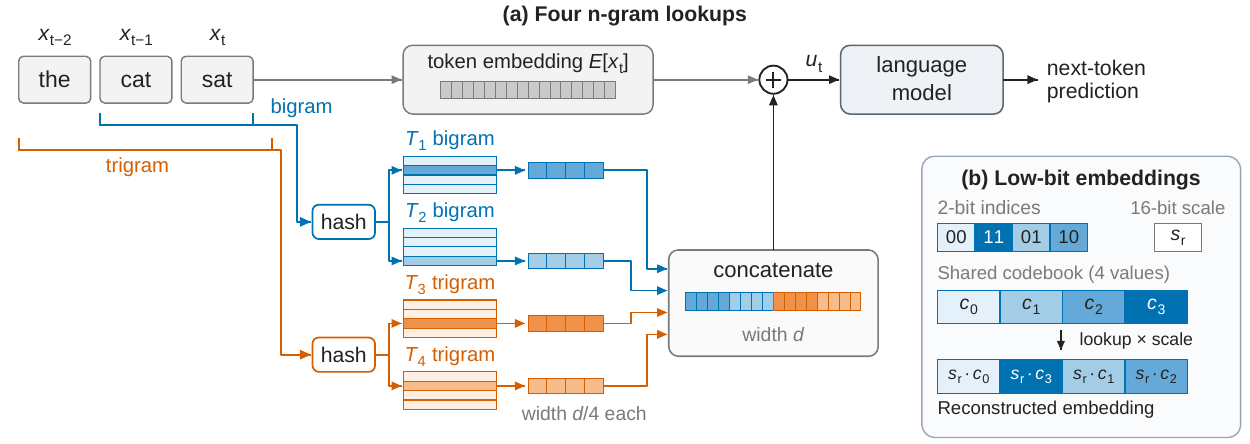}
  \caption{\textbf{Four n-gram lookups, stored at low precision.} \textbf{(a)} The bigram and trigram ending at the current token each retrieve two embeddings. The four embeddings are concatenated and added to the token embedding to form $u_t$. \textbf{(b)} Each two-bit index selects one of four codebook values. Multiplying each selected value by the embedding's 16-bit scale reconstructs the embedding. Codebook values use 16 bits and are shared across all embeddings of a table.}
  \label{fig:method}
\end{figure}

Our main contributions are:
\begin{itemize}
\item \textbf{Substantial table memory savings with little loss of performance.} Across three model architectures, three-bit n-gram tables save 80\% of table memory while retaining approximately 97\% of the validation-loss improvement from adding the tables. We also demonstrate substantial table memory savings in DeepSeek-V4.1-Flash and Qwen3.8-Flash-Next.
\item \textbf{Practical guidance on quantization and precision allocation.} We systematically compare codebook construction, scale fitting, and precision allocation across bigram and trigram tables.
\item \textbf{Performance recovery through fine-tuning.} A short fine-tuning phase substantially reduces the performance loss from low-bit compression, making aggressive compression more effective.
\item \textbf{Better performance by reinvesting saved memory.} Tables with four times as many embeddings, stored at two bits, achieve 122--126\% of the original tables' improvement while still using 46--48\% less table memory. This shows that reducing precision can enable larger, more effective n-gram memories.
\end{itemize}

%% file: sections/method.tex
\section{Method}
\label{sec:method}

\paragraph{N-gram embedding tables.} Following the Over-Tokenized Transformer~\citep{huang2025overtokenized}, we use four embedding tables: two for bigrams and two for trigrams, as illustrated in \cref{fig:method}. The embeddings in these tables are learned jointly with the language model, with one embedding stored in each row. At each token position, the bigram and trigram ending at the current token each retrieve one embedding from each of their two corresponding tables. These four embeddings are then concatenated and added to the token embedding. Previous work has shown that this method improves language modeling performance, with further gains as the number of embeddings increases~\citep{huang2025overtokenized}.

Formally, let $x_t$ denote the current token, $E$ the token embedding matrix with embedding dimension $d$, and $T_1,\ldots,T_4$ the four tables. We construct the model input as
\begin{equation}
  u_t = E[x_t] + \operatorname{Concat}_{j=1}^{4}
  T_j\!\left[h_j(x_{t-n_j+1},\ldots,x_t)\right],
  \label{eq:table}
\end{equation}
where $n_1=n_2=2$ for the bigram tables and $n_3=n_4=3$ for the trigram tables. Here, the hash function $h_j$ determines which embedding is retrieved by mapping the token sequence to an address in table $T_j$. These retrieved embeddings each have dimension $d/4$, so concatenating all four gives the same dimension as $E[x_t]$.

\paragraph{Low-bit embeddings.} We propose reducing the memory cost of these trained tables by storing their embeddings at low precision. To do this, we give each table a set of up to $2^b$ representative values, commonly called a \emph{codebook}, and represent each embedding component by a $b$-bit index into these shared codebook values. To account for differences in magnitude between embeddings, we also store one 16-bit scale per embedding. During reconstruction, the index selects a codebook value, which is then multiplied by the embedding's scale. Thus, the original component $w_{ri}$ at position $i$ of embedding $r$ is reconstructed as
\begin{equation}
  \hat w_{ri}=s_r c_{q_{ri}}, \qquad q_{ri}\in\{0,\ldots,2^b-1\},
  \label{eq:reconstruction}
\end{equation}
where $s_r$ is the scale, $c_k$ is codebook value $k$, and $q_{ri}$ is the stored index (\cref{fig:method}b).

This low-bit representation substantially reduces table memory. For an embedding with $p$ components, memory use falls from $16p$ bits to $pb+16$ bits. For instance, with $p=128$, two-bit precision reduces the memory required per embedding by approximately 87\%, from 2,048 to 272 bits.

However, bigram and trigram embeddings represent different context patterns, so they may need different precision to preserve their contribution to model performance. We therefore propose using a mixture of bit widths across the two n-gram orders, allocating more bits to the tables that are more sensitive to quantization within a fixed table memory budget.

\paragraph{Choosing the codebook values.} We consider three codebook families. First, uniform codebooks use equally spaced values. For instance, at two bits, the codebook values are $\{-1,-1/3,1/3,1\}$. We also test a variant that includes zero, using $\{-1,0,1\}$, so that zero can be represented exactly. Second, normal-quantile codebooks take values from quantiles of a standard normal distribution, placing more of them near zero. Third, we learn the codebook values from each table's trained embeddings using \emph{Lloyd--Max quantization}~\citep{lloyd1982least,max1960quantizing}.

To learn the Lloyd--Max codebook, we first normalize each embedding by its root mean square, allowing embeddings of different magnitudes to share a codebook. We initialize the codebook values at quantiles of the normalized components, then repeatedly assign components to their nearest codebook value and replace each value with the mean of its assigned components. This fitting runs once during compression, using only the trained embeddings. Appendix~\ref{app:quantization} gives the implementation details.

\paragraph{Choosing the embedding scales.} Once the codebook values are fixed, we consider three ways to choose each embedding's scale $s_r$. First, maximum-based scaling matches the largest absolute codebook value to the embedding's largest absolute component. Second, root-mean-square (RMS) scaling uses the embedding's overall magnitude, calculated as $s_r=\sqrt{\frac{1}{p}\sum_i w_{ri}^{2}}$. Third, error-fitted scaling chooses the scale to minimize the mean squared error between the original embedding and its reconstruction. We compare these choices to examine how scale selection affects compression quality.

After determining the codebook values and embedding scales, we store them together with the low-bit indices in place of the original embedding tables. During inference, the model retrieves embeddings from this stored representation without repeating the fitting procedure.

%% file: sections/results.tex
\section{Experiments}
\label{sec:experiments}

\subsection{Experimental setup}
\label{sec:setup}

\paragraph{Models and data.} We train five language models from scratch: nanochat with 26M and 50M parameters, a GPT-2-family model with 30M parameters, and Llama-family models with 31M and 322M parameters. By default, each model uses four n-gram tables, comprising two bigram and two trigram tables, with approximately 262k embeddings per table. We train on ClimbMix~\citep{diao2025climb} and repeat each experiment with three random seeds. Appendix~\ref{app:recipes} gives the model configurations and training settings.

\paragraph{Evaluation.} For each model, we first train two versions using the same experimental setup, one with full-precision (16-bit) n-gram tables and one without tables. We evaluate both on a held-out validation dataset, and their difference in validation loss measures the improvement from adding the tables. We then reduce the precision of the trained tables and evaluate the model again to measure how much of this improvement remains. We report this as the percentage of the original improvement retained, where 100\% matches the full-precision tables and 0\% matches the model trained without tables. Alongside this performance measure, we report the percentage of table memory saved. Results are averaged over three seeds, with standard deviations where shown.

\subsection{Table memory savings and model performance}
\label{sec:tradeoff}

We first ask how much table memory can be removed before the model loses the benefit of n-gram embeddings. We compress the trained tables to one through four bits using Lloyd--Max codebooks with fitted scales. To test whether larger tables can tolerate the same compression, we also increase the number of embeddings per table from the default 262k to 1M in the 26M nanochat model and both Llama-family models.

\begin{figure}[!htb]
  \centering
  \includegraphics[width=\linewidth]{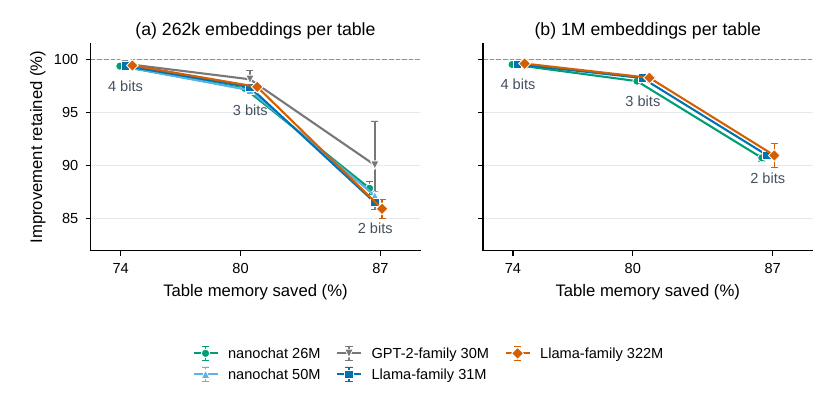}
  \caption{\textbf{Four-bit embeddings save about 75\% of table memory while retaining more than 99\% of the improvement.} (a) The five models with the default 262k embeddings per table. (b) The 26M nanochat model and both Llama-family models with 1M embeddings per table. Points and error bars show means and sample standard deviations over three seeds. The dashed line matches the original 16-bit tables.}
  \label{fig:tradeoff}
\end{figure}

The results are shown in Figure~\ref{fig:tradeoff}. We find that low-bit compression retains most of the improvement at three or four bits. For example, four-bit tables save 74--75\% of table memory while retaining more than 99\% of the original improvement. However, reducing precision further hurts performance. Two-bit tables save 86--87\% of table memory but retain only 85.9--91.0\% of the original improvement. Appendix~\ref{app:precision} gives the full results, including one-bit compression.

\subsection{Codebook choices and mixed precision}
\label{sec:design}

\paragraph{Codebook values and embedding scales.} To understand how best to choose codebook values and embedding scales, we compare the three codebook families and scale choices described in Section~\ref{sec:method}. We compare these choices using two-bit tables in the 31M Llama-family model.

\begin{figure}[!htb]
  \centering
  \includegraphics[width=\linewidth]{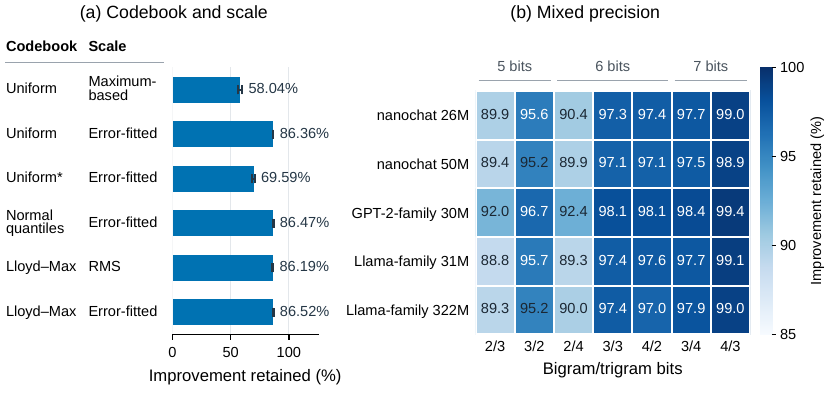}
  \caption{\textbf{Choosing an appropriate embedding scale is more important than choosing among uniform, normal-quantile, and Lloyd--Max codebooks.} (a) Two-bit comparisons on the 31M Llama-family model with 262k embeddings per table. Uniform* is the three-value variant $\{-1,0,1\}$ that includes zero. Error bars show sample standard deviations over three seeds. (b) Selected bigram/trigram precision pairs for all five models with the default 262k embeddings per table. Appendix~\ref{app:mixed} gives all precision pairs.}
  \label{fig:design}
\end{figure}

The results are shown in Figure~\ref{fig:design}a. We observe that choosing an appropriate embedding scale matters more than the choice of codebook. With error-fitted scales, the three codebooks retain a similar share of the tables' improvement (86.36--86.52\%), whereas uniform codebooks with maximum-based scaling retain only 58.04\%. Lloyd--Max also performs well with simple RMS scaling, retaining 86.19\%. Among the three fitted codebooks, Lloyd--Max provides a small additional gain, so we use it as the default in the remaining experiments.

\paragraph{Mixed precision.} Since bigram and trigram tables may require different precision, we also test all nine combinations of two-, three-, and four-bit precision. Combinations with the same total bit count have approximately equal memory cost, allowing us to compare how precision should be distributed between bigram and trigram tables. Giving more bits to bigrams consistently retains more of the improvement than the reverse allocation (Figure~\ref{fig:design}b). In particular, four-bit bigrams and three-bit trigrams retain 98.9--99.4\% of the improvement while saving 77--78\% of table memory across the five models. Appendix~\ref{app:precision} gives the full results for all precision combinations.

\FloatBarrier
\subsection{Evaluation on large existing models}
\label{sec:large}

We next test whether low-bit compression can reduce the memory used by n-gram embedding tables in large open-weight language models. We evaluate DeepSeek-V4.1-Flash~\citep{deepseek2026v41flash} and Qwen3.8-Flash-Next~\citep{qwen2026flashnext}, compressing their n-gram tables while keeping the rest of each model unchanged.

\begin{figure}[!htb]
  \centering
  \includegraphics[width=\linewidth]{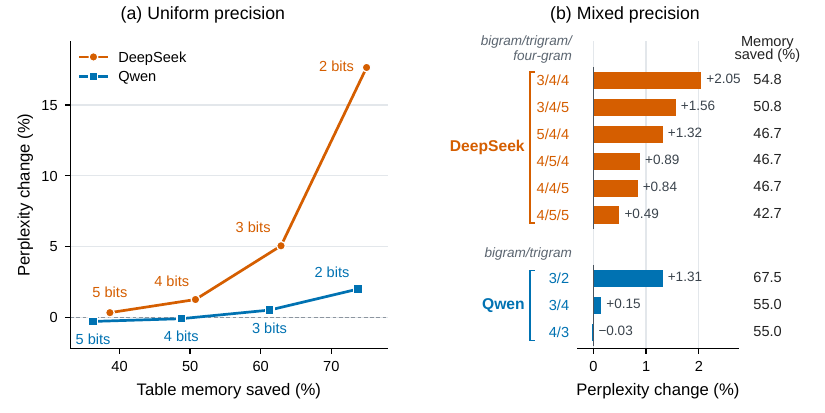}
  \caption{\textbf{Uniform and mixed precision provide different memory--performance choices in DeepSeek-V4.1-Flash and Qwen3.8-Flash-Next.} (a) Two through five bits for every table, with labels giving bits per value. (b) Representative mixed allocations, ordered by n-gram length. Both panels use WikiText-2 validation perplexity.}
  \label{fig:large}
\end{figure}

\paragraph{Uniform precision.} We first test whether using one low precision for all n-gram embedding tables can save substantial table memory without noticeably changing model performance. We compare two- to five-bit tables with the original tables using perplexity on WikiText-2 validation data~\citep{merity2017pointer}. The results are shown in Figure~\ref{fig:large}a. Four-bit tables save about half of the table memory, with a 1.26\% perplexity increase for DeepSeek and almost unchanged perplexity for Qwen. Stronger compression affects the models differently: at three bits, Qwen's perplexity increases by only 0.52\%, compared with 5.06\% for DeepSeek.

\paragraph{Mixed precision.} To see whether some n-gram tables need more precision than others, we test all combinations of two through five bits across the n-gram orders. Figure~\ref{fig:large}b shows that the precision assignment matters even at the same memory cost. For DeepSeek, giving an extra bit to trigrams or four-grams (4/5/4 or 4/4/5) preserves perplexity better than giving it to bigrams (5/4/4). In Qwen, the extra bit is more useful for bigrams: 4/3 performs better than 3/4. The better assignment therefore differs between the two models.

Besides WikiText-2, we further evaluate DeepSeek with 4/5/4-bit tables and Qwen with 4/3-bit tables on Penn Treebank and C4~\citep{marcus1993building,raffel2020exploring}. Across all three held-out datasets, these settings keep perplexity changes to less than 1.2\% while saving 47--55\% of table memory.

\begin{figure}[!htb]
  \centering
  \includegraphics[width=0.90\linewidth]{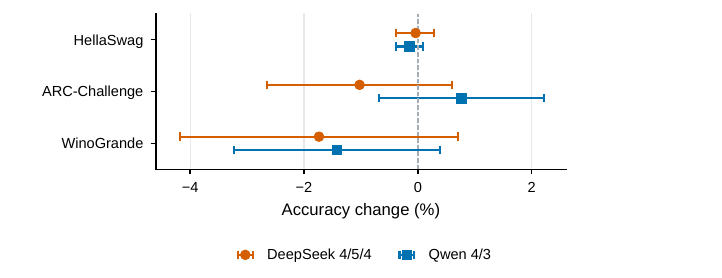}
  \caption{\textbf{Downstream evaluation of the selected mixed-precision settings.} Change in accuracy (\%) for DeepSeek 4/5/4 and Qwen 4/3, computed as low-bit accuracy minus original accuracy. Error bars show paired 95\% bootstrap confidence intervals. The settings save 46.7\% and 55.0\% of table memory, respectively.}
  \label{fig:downstream}
\end{figure}

\paragraph{Downstream performance.} In addition to perplexity, which measures next-token prediction, we examine how low-bit tables affect downstream task performance. We evaluate DeepSeek with 4/5/4-bit tables and Qwen with 4/3-bit tables on HellaSwag, ARC-Challenge, and WinoGrande~\citep{zellers2019hellaswag,clark2018arc,sakaguchi2020winogrande}. Average accuracy across these tasks decreases by only 0.93\% for DeepSeek and 0.27\% for Qwen (Figure~\ref{fig:downstream}). These results show that the table memory savings also come with little change in downstream performance.

\subsection{Recovering performance with low-bit training}
\label{sec:timing}

Compressing the n-gram tables changes the embeddings that the model learned to use, which can reduce performance. We therefore ask whether a short fine-tuning phase can recover this loss, or whether using low-bit embeddings throughout pre-training is more effective. To train with low-bit embeddings, we maintain a BF16 copy for gradient updates but quantize the retrieved embeddings during each forward pass. Both the embeddings and the rest of the model remain trainable, allowing the model to adapt to the chosen precision. We compare four sequences: (1) compress $\rightarrow$ pre-train, (2) pre-train $\rightarrow$ compress $\rightarrow$ fine-tune, (3) pre-train $\rightarrow$ fine-tune $\rightarrow$ compress, and (4) pre-train $\rightarrow$ compress. Comparing fine-tuning before and after compression tests whether recovery comes from adapting to low-bit embeddings rather than simply receiving more training.

\begin{figure}[!htb]
  \centering
  \includegraphics[width=0.90\linewidth]{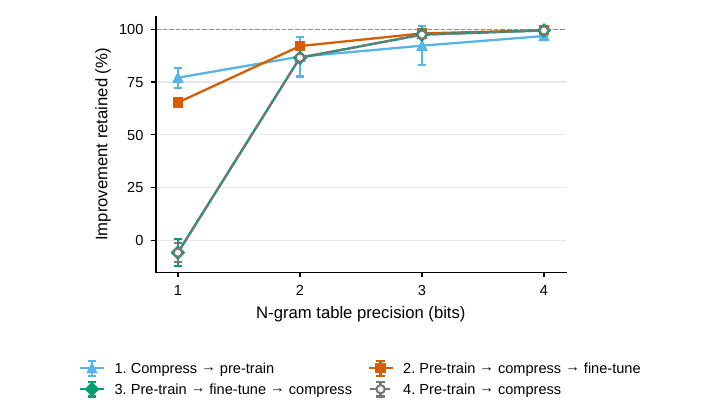}
  \caption{\textbf{Fine-tuning after compression recovers much of the n-gram improvement.} Improvement retained on the 31M Llama-family model, relative to full-precision and no-table models with the same training budget. Error bars show sample standard deviations over three seeds.}
  \label{fig:training}
\end{figure}

The results are summarized in Figure~\ref{fig:training}. We observe that a brief fine-tuning phase after compression recovers much of the performance lost at very low precision. After fine-tuning, one- and two-bit tables retain 65.3\% and 92.0\% of the improvement provided by full-precision tables, respectively. Surprisingly, one-bit embeddings perform even better when used throughout pre-training, retaining 77.0\% of the full-precision tables' improvement. This shows that even one bit per embedding component can preserve a large share of the benefit from n-gram tables when the model learns to use them from the start. Appendix~\ref{app:qat} gives the full results, while Appendix~\ref{app:hashing} examines how alternative hashing methods affect performance with low-bit embeddings.

\subsection{Reinvesting memory in larger tables}
\label{sec:rows}

\begin{figure}[!htb]
  \centering
  \includegraphics[width=\linewidth]{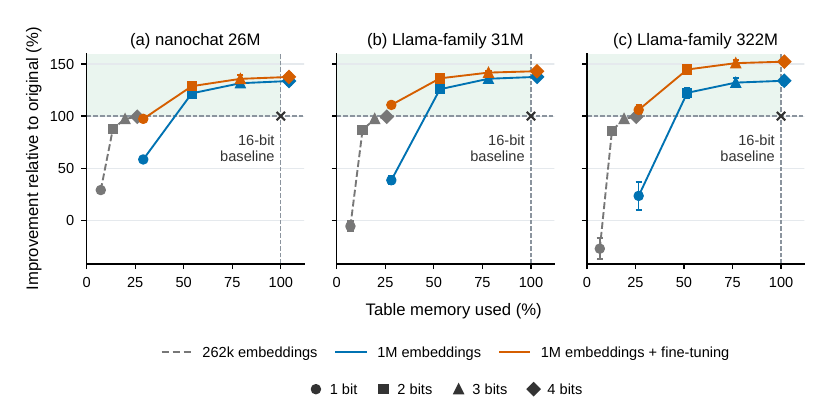}
  \caption{\textbf{Larger low-bit tables can improve performance while using less memory.} Both axes use the original 262k-embedding, 16-bit tables as the reference, marked at (100\%, 100\%). The shaded region indicates greater improvement while using less memory. Fine-tuning follows compression. Error bars show sample standard deviations over three seeds.}
  \label{fig:rows}
\end{figure}

Finally, we ask whether the memory saved by compression can support larger tables and improve performance. We use tables with 262k embeddings stored at 16-bit precision as the baseline. We then increase each table to 1M embeddings and test lower precisions to see whether the larger tables can improve performance while using less memory than the baseline. We measure performance relative to the improvement that the baseline tables provide over the model without n-gram embeddings.

The results are shown in Figure~\ref{fig:rows}. We observe that reinvesting the memory saved by compression in more embeddings can improve performance while still reducing total table memory. For example, increasing each table from 262k to 1M embeddings at two-bit precision uses only 52--54\% of the original table memory while providing 22--26\% more improvement than the original 16-bit tables. Three-bit tables improve performance further while using 77--79\% of the original table memory.

We also test a brief fine-tuning phase on the compressed larger tables, following Section~\ref{sec:timing}, to assess whether adaptation brings further improvement. Fine-tuning improves performance across all three models, with the largest recovery at one bit: with the same fine-tuning budget, compressing before fine-tuning achieves 97--111\% of the original tables' improvement, compared with 44--63\% when compressing after fine-tuning. At two bits, fine-tuning raises the improvement from 122--126\% to 129--145\% at the same memory cost. These results suggest a practical approach to memory-efficient deployment: pre-train with more embeddings, then compress the tables and briefly fine-tune the model to recover performance.

%% file: sections/related.tex
\section{Related work}
\label{sec:related}

\paragraph{N-gram and lookup memory in language models.} Learned local-context representations go back to neural n-gram models~\citep{bengio2003neural}, and hashing shares parameters among many keys~\citep{chen2015hashing,svenstrup2017hash}. N-Grammer adds n-gram embeddings over discrete latent codes~\citep{roy2022ngrammer}, and BLT uses hashed n-gram embeddings over bytes~\citep{pagnoni2025blt}. The Over-Tokenized Transformer~\citep{huang2025overtokenized} hashes multi-gram inputs into large input embedding tables, the operation we study. Deep Sparse Embedding (DSE)~\citep{cheng2026dse} places hashed n-gram memory inside the network with context-aware gating and short convolutions, balances it against mixture-of-experts capacity, and serves it from host memory. Other designs factorize n-gram embeddings~\citep{zhou2026tngram,yang2026factorengram} or use frequency-aware hashing and alias mixing~\citep{chen2026xgram}. SCONE precomputes contextualized n-gram embeddings and stores them off-accelerator~\citep{yu2025scone}. Memory layers retrieve from a product-key memory~\citep{lample2019product,berges2025memory}, and external datastores add capacity at inference time~\citep{khandelwal2020knnlm,borgeaud2022retro}.

\paragraph{Compressing embedding tables.} Large embedding tables in recommendation systems are compressed by per-embedding four-bit quantization with codebooks~\citep{guan2019embedding}, compositional embeddings~\citep{shi2020compositional}, and mixed embedding dimensions~\citep{ginart2021mixed}; word embeddings can be replaced by learned compositional codes~\citep{shu2018compositional}. Because hashed embeddings are already shared, we compare fewer embeddings with fewer bits per embedding directly.

%% file: sections/conclusion.tex
\section{Conclusion}
\label{sec:conclusion}

We propose low-bit precision for n-gram embedding tables and systematically study its effects on memory use and model performance. Across models trained from scratch, four-bit tables save approximately 75\% of table memory while preserving more than 99\% of the improvement provided by full-precision tables. Choosing appropriate embedding scales matters more than the choice of codebook, and mixed precision allows different n-gram tables to use the precision they need. At lower precision, a brief fine-tuning phase can recover much of the lost performance. These findings also extend to DeepSeek-V4.1-Flash and Qwen3.8-Flash-Next, where compression saves 47--55\% of table memory with little change in perplexity and downstream performance. Beyond reducing memory use, reinvesting the savings in more embeddings can improve performance over the original full-precision tables. Together, our results offer practical guidance for making n-gram memory more efficient and using the saved capacity to build better language models.

%% file: sections/ai_use.tex
\section{AI Use}
\label{sec:ai_use}

This work builds on a finding made by an autonomous agent run by the authors on rStation, a multi-agent research platform. Given the open-ended task of improving language-model pre-training, the agent found that n-gram embedding tables could be stored at low precision with minimal performance loss. The agent published its findings at \url{https://rstation.ai/problems/general_pretraining/papers/6/v4}.

The authors considered this finding potentially useful to the broader research community and therefore reproduced and extended the experiments outside rStation, using additional models and datasets. The authors also revised the manuscript to make the findings accessible to researchers.

%% file: sections/appendix.tex
\raggedbottom
\setcounter{topnumber}{5}
\setcounter{totalnumber}{6}
\renewcommand{\topfraction}{0.95}
\renewcommand{\textfraction}{0.05}
\interlinepenalty=1000
\FloatBarrier
\section{Detailed methods}
\label{app:methods}

\FloatBarrier
\subsection{N-gram hashing and lookup}

Following the Over-Tokenized Transformer~\citep{huang2025overtokenized}, we retrieve embeddings by hashing the bigram and trigram ending at each token. For the token ID $x_t$ at position $t$, vocabulary size $V=8192$, and base $B=V+1$, our from-scratch experiments use the keys
\begin{equation}
 k_t^{(2)}=x_t+Bx_{t-1},\qquad k_t^{(3)}=x_t+Bx_{t-1}+B^2x_{t-2}.
\end{equation}
Missing predecessors take the value $V$. The bigram addresses are $k_t^{(2)}\bmod m$ and $k_t^{(2)}\bmod(m+2)$, and the trigram addresses are $k_t^{(3)}\bmod(m+4)$ and $k_t^{(3)}\bmod(m+6)$. The base table size is $m=262{,}141$ by default and $m=1{,}048{,}573$ for the 1M-embedding experiments. The retrieved embeddings are then used as described in Section~\ref{sec:method}.

\FloatBarrier
\subsection{Codebook construction}
\label{app:codebook_construction}
\label{app:quantization}

As described in Section~\ref{sec:method}, we consider three codebook families: uniform, normal-quantile, and Lloyd--Max. Each codebook supplies the representative values shared by the embeddings in one table. With $b$ bits per component, a codebook can contain up to $K=2^b$ values, denoted $c_0,\ldots,c_{K-1}$.

\paragraph{Uniform codebooks.} We place $K$ values at equal intervals between $-1$ and $1$. At two bits, this gives $\{-1,-1/3,1/3,1\}$. We also test the zero-centered variant, denoted Uniform*, which uses $K-1$ equally spaced values so that zero is included. Its two-bit codebook is $\{-1,0,1\}$.

\paragraph{Normal-quantile codebooks.} We place values at quantiles of a standard normal distribution, concentrating more values near zero. For more than two bits, we take quantiles at probabilities $(k+1/2)/K$ for $k=0,\ldots,K-1$, then divide all values by their largest absolute value. At two bits, we instead use the fixed codebook $\{-1,-0.3186,0.3186,1\}$.

\paragraph{Lloyd--Max codebooks.} To adapt the codebook to the trained embeddings, we learn its values from their component distribution~\citep{lloyd1982least,max1960quantizing}. We first normalize each embedding so that differences in magnitude do not dominate the shared codebook. Let $w_{ri}$ be component $i$ of embedding $r$, and let $p$ be its number of components. The initial normalization scale $s_r^{(0)}$ is the embedding's root mean square, and $z_{ri}$ is its normalized component:
\begin{equation}
 s_r^{(0)}=\sqrt{\frac{1}{p}\sum_{i=1}^{p}w_{ri}^{2}},\qquad
 z_{ri}=\frac{w_{ri}}{s_r^{(0)}}.
 \label{eq:lloyd}
\end{equation}
In implementation, we clamp the normalization scale to at least $10^{-12}$ to avoid division by zero.

We then fit the codebook to these normalized components. We sample at most 4M components per table and initialize the $K$ codebook values at quantiles $(k+1/2)/K$ of this sample. Each Lloyd--Max iteration has two steps. First, we assign every sampled component to its nearest codebook value. Second, we replace each codebook value with the mean of the components assigned to it. Formally, for sampled components indexed by $j$, the assignment $a_j$ and updated value $c_k^{\mathrm{new}}$ are
\begin{equation}
 a_j=\operatorname*{arg\,min}_{0\leq k<K}|z_j-c_k|^2,\qquad
 c_k^{\mathrm{new}}=\frac{1}{|\mathcal S_k|}\sum_{j\in\mathcal S_k}z_j,
 \quad \mathcal S_k=\{j:a_j=k\}.
\end{equation}
Here, $\mathcal S_k$ contains the indices of the sampled components assigned to value $k$. If this set is empty, that codebook value is unchanged. We repeat the two steps 60 times and store the fitted values in bfloat16 (BF16). Once the codebook is fixed, we choose the embedding scales using one of the methods below.

\FloatBarrier
\subsection{Embedding scales}
\label{app:scales}

As described in Section~\ref{sec:method}, we compare three methods for choosing embedding scales: maximum-based, root-mean-square (RMS), and error-fitted scaling. For embedding $r$, the scale $s_r$ multiplies every selected codebook value, so component $w_{ri}$ is reconstructed as $s_r c_{q_{ri}}$, where $q_{ri}$ is the index of the selected codebook value. The methods below differ in how they choose $s_r$ for a fixed codebook.

\paragraph{Maximum-based scaling.} We match the largest absolute codebook value to the embedding's largest absolute component. With codebook values $c_k$, the scale is
\begin{equation}
 s_r^{\mathrm{max}}=\frac{\max_i|w_{ri}|}{\max_k|c_k|}.
\end{equation}

\paragraph{Root-mean-square (RMS) scaling.} We use the embedding's overall magnitude as the scale, i.e.
\begin{equation}
 s_r^{\mathrm{RMS}}=\sqrt{\frac{1}{p}\sum_{i=1}^{p}w_{ri}^{2}}.
\end{equation}

\paragraph{Error-fitted scaling.} To choose a scale based on reconstruction quality, we test the RMS scale and a range of scales derived from maximum-based scaling. Specifically, the candidates are $s_r^{\mathrm{RMS}}$ and $\alpha s_r^{\mathrm{max}}$, where $\alpha$ is a multiplier taking values $1.00,0.95,\ldots,0.30$. A smaller $\alpha$ brings the representative values closer together, allowing finer approximation of smaller components while clipping larger ones. For each candidate, we round the scale to BF16, assign each component to its nearest scaled codebook value, and reconstruct the embedding in BF16. We select the candidate with the lowest mean squared error relative to the original embedding. This is the error-fitted method used in the main compression comparisons.

\FloatBarrier
\section{Experimental setup}
\label{app:setup}

\FloatBarrier
\subsection{Models and training settings}
\label{app:recipes}

We train five models from scratch to test compression across architectures and model sizes. Table~\ref{tab:configs} lists their dimensions and n-gram table sizes. The two nanochat models follow \citet{karpathy2026autoresearch,karpathy2025nanochat}, while the other models use GPT-2-family~\citep{radford2019gpt2} and Llama-family architectures, with the latter adapted from SmolLM2~\citep{allal2025smollm2}. We refer to these implementations for the remaining architectural details. All models use an 8,192-token vocabulary and the same n-gram lookup design. The source code underlying the experiments can be found at \url{https://github.com/dualverse-ai/lowbit-ngram}.

\begin{table}[!htbp]
  \centering
  \caption{Model dimensions and n-gram table sizes. Parameter counts exclude the n-gram tables, and $d$ denotes model width.}
  \label{tab:configs}
\begin{dualversetable}
  \begin{tabular}{l>{\DualverseNumT{000.00}}c>{\DualverseNumT{00}}c>{\DualverseNumT{000}}c>{\DualverseNumT{262k, 1M}}c}
    \toprule
    \DualverseHeadRow
    \DualverseHead[l]{Model} & \DualverseHead{Parameters\\(millions)} & \DualverseHead{Layers} & \DualverseHead{Width $d$} & \DualverseHead{Embeddings\\per table} \\
    \midrule
    nanochat 26M & \DualverseNum{26.35} & \DualverseNum{6} & \DualverseNum{384} & \DualverseNumL{262k, 1M} \\
    nanochat 50M & \DualverseNum{50.33} & \DualverseNum{8} & \DualverseNum{512} & \DualverseNumL{262k} \\
    GPT-2-family 30M & \DualverseNum{30.46} & \DualverseNum{8} & \DualverseNum{512} & \DualverseNumL{262k} \\
    Llama-family 31M & \DualverseNum{30.94} & \DualverseNum{8} & \DualverseNum{512} & \DualverseNumL{262k, 1M} \\
    Llama-family 322M & \DualverseNum{322.50} & \DualverseNum{32} & \DualverseNum{960} & \DualverseNumL{262k, 1M} \\
    \bottomrule
  \end{tabular}
\end{dualversetable}
\end{table}

Table~\ref{tab:training_settings} lists the optimizers, peak learning rates, and training budgets. Models with and without tables use the same training setup, and the table embeddings are initialized to zero.

\begin{table}[!htbp]
  \centering
  \caption{Training settings. For nanochat, the backbone learning rate applies to Muon-updated matrices. Other parameter groups follow the original implementation.}
  \label{tab:training_settings}
\begin{dualversetable}
  \begin{tabular}{l*{3}{C{2.55cm}}}
    \toprule
    \DualverseHeadRow
    \DualverseHead[l]{Setting} & \DualverseHead{nanochat} & \DualverseHead{GPT-2 30M /\\Llama 31M} & \DualverseHead{Llama 322M} \\
    \midrule
    \DualverseTableSection{4}{Optimization}
    \DualverseIndent Backbone optimizer & Muon / AdamW & AdamW & AdamW \\
    \DualverseIndent Table optimizer & AdamW & AdamW & AdamW \\
    \DualverseIndent Backbone learning rate & 0.04 & 0.001 & 0.0006 \\
    \DualverseIndent Table learning rate & $0.6\sqrt{768/d}$ & 0.003 & 0.0018 \\
    \addlinespace[2pt]
    \DualverseTableSection{4}{Training budget}
    \DualverseIndent Updates & 953 & 7,624 & 7,630 \\
    \DualverseIndent Target tokens per update & 524,288 & 65,536 & 131,072 \\
    \bottomrule
  \end{tabular}
\end{dualversetable}
\end{table}

\subsection{Training and validation data}
\label{app:data}

Training uses the shuffled ClimbMix release~\citep{diao2025climb}, tokenized with an 8,192-token byte-pair encoding tokenizer. We pack documents into sequences of 2,049 tokens, providing 2,048 prediction targets per sequence. The small models train on approximately 500M prediction targets, and the 322M model trains on approximately 1B prediction targets.

We use separate validation datasets to measure final performance. These datasets come from held-out portions of ClimbMix. We remove duplicate validation documents and any documents also present in training or development. The validation set contains 10,240 packed sequences, totaling 20.97M prediction targets. All compared models are scored on the same sequences.

\FloatBarrier
\subsection{DeepSeek and Qwen evaluation}
\label{app:large_setup}

For DeepSeek-V4.1-Flash and Qwen3.8-Flash-Next, we compress the n-gram tables while keeping the rest of each model fixed. Their original tables are stored at eight-bit floating-point (FP8) precision, which we use as the baseline when measuring table memory savings. We use WikiText-2 validation data to compare precision allocations. DeepSeek has bigram, trigram, and four-gram tables, while Qwen has bigram and trigram tables. For both models, we use Lloyd--Max codebooks with error-fitted embedding scales as the default for evaluation.

To test whether promising allocations transfer beyond the selection data, we evaluate them on held-out WikiText-2, Penn Treebank, and C4. Each corpus contributes 64 windows and 32,768 scored target tokens, with identical windows for the original and compressed models.

We evaluate the models on HellaSwag, ARC-Challenge, and WinoGrande. HellaSwag and ARC-Challenge use length-normalized accuracy, while WinoGrande uses accuracy. We evaluate their full splits of 10,042, 1,172, and 1,267 examples, respectively. Accuracy differences compare predictions on the same examples, with 95\% confidence intervals from 100,000 paired bootstrap samples.

\FloatBarrier
\subsection{Training and fine-tuning with low-bit embeddings}
\label{app:qat_setup}

This section describes the training and fine-tuning procedures used in Sections~\ref{sec:timing} and~\ref{sec:rows}.

\paragraph{Training with low-bit embeddings.} We retain a BF16 copy of each embedding for gradient updates and quantize the retrieved embeddings in the forward pass. During backpropagation, we treat quantization as an identity operation, passing gradients to the BF16 copy through a straight-through estimator. Embedding scales are fitted on each forward pass, and the Lloyd--Max codebook is refitted every update for the first 100 updates and every 25 updates thereafter.

\paragraph{Fine-tuning details.} For the 262k-embedding tables, fine-tuning uses 381 additional updates on unseen training sequences, with 32 sequences per update. All weights remain trainable, using fresh AdamW states, 20 warmup updates, and a learning rate of 0.1 times the original peak.

For the 1M-embedding experiments, fine-tuning uses new training sequences for the smaller models and reuses pre-training sequences for the 322M model.

\FloatBarrier
\subsection{Computational resources}
\label{app:repro}

The from-scratch experiments run on RTX 5090 GPUs. Small-model training runs take approximately 10--25 minutes on one GPU, while a 322M run takes 6.4--6.8 hours. DeepSeek and Qwen evaluations use four H200 GPUs with tensor parallelism of four.

\paragraph{Table memory.} We compute table memory savings from the memory required by the n-gram embedding tables. The compressed table memory includes the low-bit embedding indices, one 16-bit scale per embedding, and the shared codebook values. We report the percentage reduction relative to the original tables, stored at 16 bits in our from-scratch experiments and eight bits in DeepSeek and Qwen. These percentages describe table memory rather than total model memory.

\clearpage
\section{Complete experimental results}
\label{app:precision}

\FloatBarrier
\subsection{Compression and table size}
\label{app:size}

Here we present the full compression results for Section~\ref{sec:tradeoff} (Figure~\ref{fig:tradeoff}) and the table-size comparison in Section~\ref{sec:rows} (Figure~\ref{fig:rows}). Tables~\ref{tab:retention} and~\ref{tab:size} report results for 262k and 1M embeddings per table, respectively. Both use each model's 262k-embedding, 16-bit tables as the reference for computing Loss Change and Improvement.

\begin{table}[!htbp]
\centering
\caption{Compression with 262k embeddings per table, using Lloyd--Max codebooks and error-fitted scales. Validation loss, loss change, and improvement are means $\pm$ sample standard deviations over three seeds.}
\label{tab:retention}
\begin{dualversetable}
\input{sections/generated/tab_retention}
\end{dualversetable}
\end{table}

\begin{table}[!htbp]
\centering
\caption{Compression with 1M embeddings per table, using Lloyd--Max codebooks and error-fitted scales. Validation loss, loss change, and improvement are means $\pm$ sample standard deviations over three seeds.}
\label{tab:size}
\begin{dualversetable}
\input{sections/generated/tab_rows}
\end{dualversetable}
\end{table}

\clearpage
\subsection{Codebook and scale choices}
\label{app:codebooks}

Here we present the full codebook and scale comparisons for Section~\ref{sec:design} (Figure~\ref{fig:design}a). Table~\ref{tab:design} extends the comparison to the other models and includes the uniform codebook variants.

\begin{table}[!htbp]
\centering
\caption{Validation loss (BPB), averaged over three seeds. Uniform* includes zero and uses $2^b-1$ values.}
\label{tab:design}
\begin{dualversetable}
\input{sections/generated/tab_design}
\end{dualversetable}
\end{table}

\clearpage
\subsection{Mixed precision}
\label{app:mixed}

\begin{figure}[!htbp]
\centering
\includegraphics[width=\linewidth]{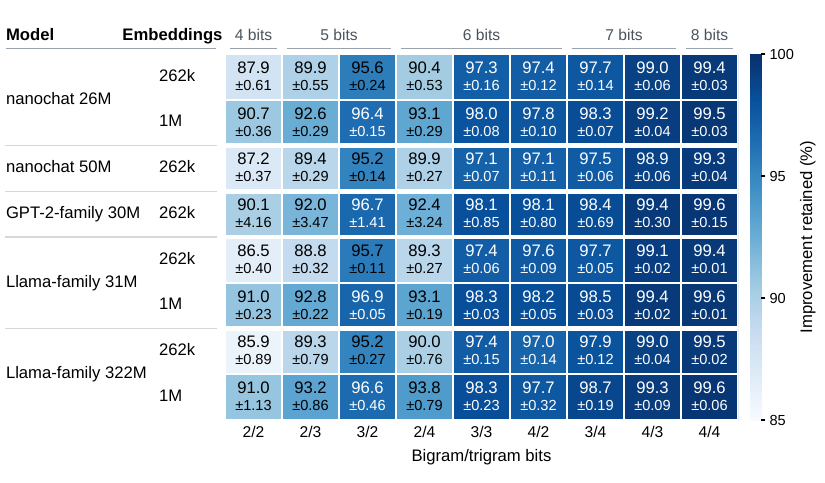}
\caption{Complete bigram/trigram precision sweep with Lloyd--Max codebooks and error-fitted scales. Entries give improvement retained (\%) as mean $\pm$ sample standard deviation over three seeds.}
\label{fig:mixedcomplete}
\end{figure}

Here we present the full mixed-precision results for Section~\ref{sec:design} (Figure~\ref{fig:design}b). Figure~\ref{fig:mixedcomplete} gives all nine bigram/trigram precision pairs drawn from two, three, and four bits, for both 262k and 1M embeddings per table. At approximately equal memory cost, assigning more bits to bigrams consistently preserves more of the improvement than the reverse allocation. This pattern holds across all five models and both table sizes tested.

\FloatBarrier
\subsection{DeepSeek and Qwen}
\label{app:large}

\begin{figure}[!htbp]
\centering
\includegraphics[width=0.90\linewidth]{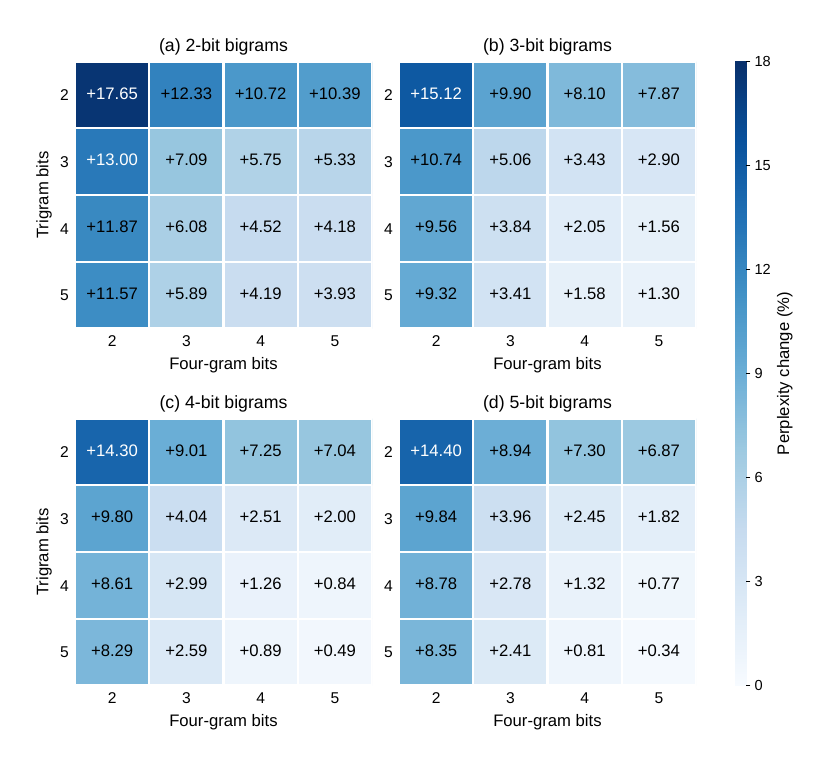}
\caption{DeepSeek-V4.1-Flash: all 64 bigram/trigram/four-gram precision allocations with Lloyd--Max codebooks and error-fitted scales. Entries show validation perplexity change (\%).}
\label{fig:deepseekgrid}
\end{figure}

\begin{figure}[!htbp]
\centering
\includegraphics[width=\linewidth]{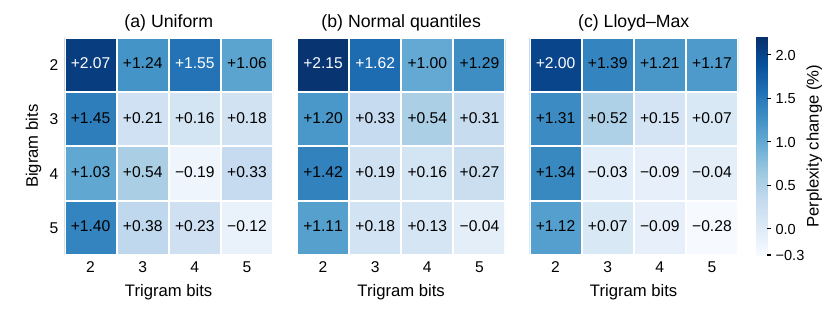}
\caption{Qwen3.8-Flash-Next: all sixteen bigram/trigram allocations for each codebook, using error-fitted scales. Entries show validation perplexity change (\%).}
\label{fig:qwengrid}
\end{figure}

Here we present the full DeepSeek and Qwen results for Section~\ref{sec:large} (Figures~\ref{fig:large} and~\ref{fig:downstream}). Figures~\ref{fig:deepseekgrid} and~\ref{fig:qwengrid} give the complete precision sweeps. Tables~\ref{tab:largeppl} and~\ref{tab:largedown} then report held-out perplexity and downstream accuracy for the 4/5/4-bit DeepSeek and 4/3-bit Qwen settings.

\FloatBarrier

\begin{table}[!htbp]
\centering
\caption{Perplexity before and after compression with DeepSeek 4/5/4-bit tables and Qwen 4/3-bit tables. Confidence intervals for perplexity change are computed by paired bootstrap resampling of evaluation windows.}
\label{tab:largeppl}
\begin{dualversetable}
\input{sections/generated/tab_large_ppl}
\end{dualversetable}
\end{table}

\begin{table}[!htbp]
\centering
\caption{Downstream accuracy (\%), with low-bit minus original accuracy and paired 95\% bootstrap confidence intervals.}
\label{tab:largedown}
\begin{dualversetable}
\input{sections/generated/tab_large_downstream}
\end{dualversetable}
\end{table}

\clearpage
\subsection{Training and fine-tuning with low-bit embeddings}
\label{app:qat}

Table~\ref{tab:timing} gives the full results for the four training sequences in Figure~\ref{fig:training}. Table~\ref{tab:rowsft} gives the larger-table fine-tuning results behind Figure~\ref{fig:rows}.

\begin{table}[!htbp]
\centering
\caption{Improvement retained (\%) for the four training sequences on the 31M Llama-family model, with 262k embeddings per table. Means and sample standard deviations over three seeds. Fine-tuning comparisons use equally fine-tuned references.}
\label{tab:timing}
\begin{dualversetable}
\input{sections/generated/tab_timing}
\end{dualversetable}
\end{table}

\begin{table}[!htbp]
\centering
\caption{Fine-tuning 1M-embedding tables before or after compression. Improvement and table memory are relative to the original 262k-embedding, 16-bit tables. Results are means and sample standard deviations over three seeds.}
\label{tab:rowsft}
\begin{dualversetable}
\input{sections/generated/tab_rows_finetune}
\end{dualversetable}
\end{table}

\FloatBarrier

\clearpage
\subsection{Alternative hashing methods}
\label{app:hashing}
\begingroup
\setlength{\intextsep}{6pt}

To examine whether the way n-grams share embedding parameters affects low-bit performance, we compare conventional whole-embedding hashing with two component-wise alternatives. Conventional hashing maps an n-gram to one embedding in each table and retrieves all its components together. Consequently, two n-grams mapped to the same embedding share every component. The alternatives instead choose a separate address for each output component while keeping the same table size and embedding dimension.

\paragraph{Same-column hashing.} We retain the table as a matrix, with one embedding per row and one component per column. To construct component $j$ of an n-gram embedding, we hash the n-gram together with $j$ to choose a row, then retrieve the value in column $j$ of that row. Different components can therefore come from different embeddings, but the first component always comes from the first column, the second from the second column, and so on.

\paragraph{Fully random component hashing.} We flatten the same table into a vector and hash the n-gram together with each component index to select a value anywhere in that vector. Unlike same-column hashing, the selected value can come from any column as well as any row. In both variants, the hash addresses are fixed: the same n-gram always retrieves the same components, whose values are learned during training.

\paragraph{Experimental setup.} We compare all three methods on nanochat 26M, GPT-2-family 30M, and Llama-family 31M, using the default 262k embeddings per table, ClimbMix, and three seeds. Each comparison uses the same training setup and evaluates both 16-bit and two-bit tables. The two-bit models use quantized embeddings throughout pre-training, following Appendix~\ref{app:qat_setup}. We quantize each table using Lloyd--Max codebooks and error-fitted scales before gathering its components, so each selected component keeps the scale of its source embedding.

\begin{table}[!htbp]
\centering
\caption{Alternative hashing methods at 16-bit and two-bit precision. Values give the improvement relative to conventional 16-bit tables (\%), reported as mean $\pm$ sample standard deviation over three seeds. For every row, 100\% denotes the validation-loss reduction from adding conventional 16-bit tables to the corresponding model without tables.}
\label{tab:hashing}
\begin{dualversetable}
\renewcommand{\arraystretch}{1.10}
\input{sections/generated/tab_hashing}
\end{dualversetable}
\end{table}

\paragraph{Discussion.} Table~\ref{tab:hashing} shows that both component-wise methods improve the average two-bit result over conventional hashing on all three architectures. Their improvement reaches 93.6--100.3\% of that provided by conventional 16-bit tables, compared with 87.0--93.7\% for conventional two-bit hashing. The gains are larger on GPT-2 and Llama than on nanochat, suggesting that the choice of shared embedding components can help preserve performance at low precision.

Same-column hashing also appears promising beyond low precision. It improves mean performance over conventional hashing in five of the six settings, including two of the three full-precision comparisons. One possible explanation is that component-wise hashing allows n-grams to share individual components without sharing an entire embedding, reducing interference from whole-embedding collisions. These results suggest that same-column hashing may offer broader benefits for language modeling. Establishing how consistently it helps and whether collision patterns explain these gains deserves further investigation beyond the scope of this study, which focuses on low-precision embeddings.
\FloatBarrier
\endgroup

%% file: sections/generated/tab_retention.tex
\begin{tabularx}{\linewidth}{%
  >{\DualverseNumT{00}\raggedright\arraybackslash}p{1.55cm}%
  >{\DualverseNumT{000.0}\centering\arraybackslash}X%
  >{\DualversePMT{0.00000}{0.00000}\centering\arraybackslash}X%
  >{\DualversePMT{$-$0.000}{0.000}\centering\arraybackslash}X%
  >{\DualversePMT{$-$00.0}{00.0}\centering\arraybackslash}X}
\toprule
\DualverseHeadRow
\DualverseHead[l]{Precision\\(bits)} & \DualverseHead{Memory used\\(\%)} & \DualverseHead{Validation loss\\(bits/byte)} & \DualverseHead{Loss change\\(\%)} & \DualverseHead{Improvement\\(\%)} \\
\midrule
\DualverseTableSection{5}{nanochat 26M}
\DualverseIndent\DualverseNum{1} & \DualverseNum{7.3} & \DualverseMSD{1.03963}{0.00032} & \DualverseMSD{2.132}{0.049} & \DualverseMSD{29.0}{2.8} \\
\DualverseIndent\DualverseNum{2} & \DualverseNum{13.5} & \DualverseMSD{1.02164}{0.00066} & \DualverseMSD{0.365}{0.004} & \DualverseMSD{87.9}{0.6} \\
\DualverseIndent\DualverseNum{3} & \DualverseNum{19.8} & \DualverseMSD{1.01876}{0.00070} & \DualverseMSD{0.082}{0.001} & \DualverseMSD{97.3}{0.2} \\
\DualverseIndent\DualverseNum{4} & \DualverseNum{26.0} & \DualverseMSD{1.01812}{0.00070} & \DualverseMSD{0.018}{0.000} & \DualverseMSD{99.4}{0.0} \\
\DualverseIndent\DualverseNum{16} & \DualverseNum{100.0} & \DualverseMSD{1.01793}{0.00070} & \DualverseMSD{0.000}{0.000} & \DualverseMSD{100.0}{0.0} \\
\DualverseIndent No table & \DualverseNum{0.0} & \DualverseMSD{1.04855}{0.00139} & \DualverseMSD{3.008}{0.178} & \DualverseMSD{0.0}{0.0} \\
\addlinespace[2pt]
\DualverseTableSection{5}{nanochat 50M}
\DualverseIndent\DualverseNum{1} & \DualverseNum{7.0} & \DualverseMSD{0.98935}{0.00093} & \DualverseMSD{1.636}{0.006} & \DualverseMSD{22.7}{2.4} \\
\DualverseIndent\DualverseNum{2} & \DualverseNum{13.3} & \DualverseMSD{0.97606}{0.00087} & \DualverseMSD{0.272}{0.002} & \DualverseMSD{87.2}{0.4} \\
\DualverseIndent\DualverseNum{3} & \DualverseNum{19.5} & \DualverseMSD{0.97402}{0.00086} & \DualverseMSD{0.061}{0.001} & \DualverseMSD{97.1}{0.1} \\
\DualverseIndent\DualverseNum{4} & \DualverseNum{25.8} & \DualverseMSD{0.97356}{0.00086} & \DualverseMSD{0.014}{0.000} & \DualverseMSD{99.3}{0.0} \\
\DualverseIndent\DualverseNum{16} & \DualverseNum{100.0} & \DualverseMSD{0.97342}{0.00086} & \DualverseMSD{0.000}{0.000} & \DualverseMSD{100.0}{0.0} \\
\DualverseIndent No table & \DualverseNum{0.0} & \DualverseMSD{0.99403}{0.00131} & \DualverseMSD{2.117}{0.070} & \DualverseMSD{0.0}{0.0} \\
\addlinespace[2pt]
\DualverseTableSection{5}{GPT-2-family 30M}
\DualverseIndent\DualverseNum{1} & \DualverseNum{7.0} & \DualverseMSD{1.12362}{0.00441} & \DualverseMSD{4.356}{0.086} & \DualverseMSD{16.9}{33.8} \\
\DualverseIndent\DualverseNum{2} & \DualverseNum{13.3} & \DualverseMSD{1.08232}{0.00383} & \DualverseMSD{0.521}{0.018} & \DualverseMSD{90.1}{4.2} \\
\DualverseIndent\DualverseNum{3} & \DualverseNum{19.5} & \DualverseMSD{1.07774}{0.00370} & \DualverseMSD{0.096}{0.005} & \DualverseMSD{98.1}{0.8} \\
\DualverseIndent\DualverseNum{4} & \DualverseNum{25.8} & \DualverseMSD{1.07693}{0.00364} & \DualverseMSD{0.021}{0.002} & \DualverseMSD{99.6}{0.2} \\
\DualverseIndent\DualverseNum{16} & \DualverseNum{100.0} & \DualverseMSD{1.07671}{0.00366} & \DualverseMSD{0.000}{0.000} & \DualverseMSD{100.0}{0.0} \\
\DualverseIndent No table & \DualverseNum{0.0} & \DualverseMSD{1.13947}{0.02199} & \DualverseMSD{5.832}{2.241} & \DualverseMSD{0.0}{0.0} \\
\addlinespace[2pt]
\DualverseTableSection{5}{Llama-family 31M}
\DualverseIndent\DualverseNum{1} & \DualverseNum{7.0} & \DualverseMSD{1.06613}{0.00042} & \DualverseMSD{2.555}{0.071} & \DualverseMSD{$-$5.7}{4.6} \\
\DualverseIndent\DualverseNum{2} & \DualverseNum{13.3} & \DualverseMSD{1.04295}{0.00024} & \DualverseMSD{0.326}{0.007} & \DualverseMSD{86.5}{0.4} \\
\DualverseIndent\DualverseNum{3} & \DualverseNum{19.5} & \DualverseMSD{1.04021}{0.00029} & \DualverseMSD{0.063}{0.002} & \DualverseMSD{97.4}{0.1} \\
\DualverseIndent\DualverseNum{4} & \DualverseNum{25.8} & \DualverseMSD{1.03971}{0.00030} & \DualverseMSD{0.014}{0.000} & \DualverseMSD{99.4}{0.0} \\
\DualverseIndent\DualverseNum{16} & \DualverseNum{100.0} & \DualverseMSD{1.03956}{0.00031} & \DualverseMSD{0.000}{0.000} & \DualverseMSD{100.0}{0.0} \\
\DualverseIndent No table & \DualverseNum{0.0} & \DualverseMSD{1.06471}{0.00083} & \DualverseMSD{2.419}{0.069} & \DualverseMSD{0.0}{0.0} \\
\addlinespace[2pt]
\DualverseTableSection{5}{Llama-family 322M}
\DualverseIndent\DualverseNum{1} & \DualverseNum{6.7} & \DualverseMSD{0.91234}{0.00070} & \DualverseMSD{1.591}{0.102} & \DualverseMSD{$-$27.3}{10.3} \\
\DualverseIndent\DualverseNum{2} & \DualverseNum{12.9} & \DualverseMSD{0.89963}{0.00040} & \DualverseMSD{0.176}{0.010} & \DualverseMSD{85.9}{0.9} \\
\DualverseIndent\DualverseNum{3} & \DualverseNum{19.2} & \DualverseMSD{0.89834}{0.00043} & \DualverseMSD{0.032}{0.001} & \DualverseMSD{97.4}{0.1} \\
\DualverseIndent\DualverseNum{4} & \DualverseNum{25.4} & \DualverseMSD{0.89811}{0.00043} & \DualverseMSD{0.007}{0.000} & \DualverseMSD{99.5}{0.0} \\
\DualverseIndent\DualverseNum{16} & \DualverseNum{100.0} & \DualverseMSD{0.89805}{0.00043} & \DualverseMSD{0.000}{0.000} & \DualverseMSD{100.0}{0.0} \\
\DualverseIndent No table & \DualverseNum{0.0} & \DualverseMSD{0.90931}{0.00111} & \DualverseMSD{1.254}{0.096} & \DualverseMSD{0.0}{0.0} \\
\bottomrule
\end{tabularx}

%% file: sections/generated/tab_rows.tex
\begin{tabularx}{\linewidth}{%
  >{\DualverseNumT{00}\raggedright\arraybackslash}p{1.55cm}%
  >{\DualverseNumT{000.0}\centering\arraybackslash}X%
  >{\DualversePMT{0.00000}{0.00000}\centering\arraybackslash}X%
  >{\DualversePMT{$-$0.000}{0.000}\centering\arraybackslash}X%
  >{\DualversePMT{$-$00.0}{00.0}\centering\arraybackslash}X}
\toprule
\DualverseHeadRow
\DualverseHead[l]{Precision\\(bits)} & \DualverseHead{Memory used\\(\%)} & \DualverseHead{Validation loss\\(bits/byte)} & \DualverseHead{Loss change\\(\%)} & \DualverseHead{Improvement\\(\%)} \\
\midrule
\DualverseTableSection{5}{nanochat 26M}
\DualverseIndent\DualverseNum{1} & \DualverseNum{29.2} & \DualverseMSD{1.03065}{0.00097} & \DualverseMSD{1.250}{0.050} & \DualverseMSD{58.4}{2.2} \\
\DualverseIndent\DualverseNum{2} & \DualverseNum{54.2} & \DualverseMSD{1.01127}{0.00023} & \DualverseMSD{$-$0.655}{0.047} & \DualverseMSD{121.9}{2.6} \\
\DualverseIndent\DualverseNum{3} & \DualverseNum{79.2} & \DualverseMSD{1.00829}{0.00021} & \DualverseMSD{$-$0.947}{0.049} & \DualverseMSD{131.6}{3.2} \\
\DualverseIndent\DualverseNum{4} & \DualverseNum{104.2} & \DualverseMSD{1.00765}{0.00021} & \DualverseMSD{$-$1.010}{0.049} & \DualverseMSD{133.7}{3.3} \\
\DualverseIndent\DualverseNum{16} & \DualverseNum{400.0} & \DualverseMSD{1.00746}{0.00021} & \DualverseMSD{$-$1.028}{0.049} & \DualverseMSD{134.3}{3.4} \\
\addlinespace[2pt]
\DualverseTableSection{5}{Llama-family 31M}
\DualverseIndent\DualverseNum{1} & \DualverseNum{28.1} & \DualverseMSD{1.05502}{0.00045} & \DualverseMSD{1.487}{0.070} & \DualverseMSD{38.5}{4.0} \\
\DualverseIndent\DualverseNum{2} & \DualverseNum{53.1} & \DualverseMSD{1.03307}{0.00042} & \DualverseMSD{$-$0.625}{0.059} & \DualverseMSD{125.8}{2.0} \\
\DualverseIndent\DualverseNum{3} & \DualverseNum{78.1} & \DualverseMSD{1.03054}{0.00044} & \DualverseMSD{$-$0.868}{0.061} & \DualverseMSD{135.9}{1.8} \\
\DualverseIndent\DualverseNum{4} & \DualverseNum{103.1} & \DualverseMSD{1.03008}{0.00045} & \DualverseMSD{$-$0.912}{0.061} & \DualverseMSD{137.7}{1.8} \\
\DualverseIndent\DualverseNum{16} & \DualverseNum{400.0} & \DualverseMSD{1.02994}{0.00045} & \DualverseMSD{$-$0.926}{0.062} & \DualverseMSD{138.3}{1.8} \\
\addlinespace[2pt]
\DualverseTableSection{5}{Llama-family 322M}
\DualverseIndent\DualverseNum{1} & \DualverseNum{26.7} & \DualverseMSD{0.90660}{0.00065} & \DualverseMSD{0.952}{0.091} & \DualverseMSD{23.4}{13.5} \\
\DualverseIndent\DualverseNum{2} & \DualverseNum{51.7} & \DualverseMSD{0.89551}{0.00037} & \DualverseMSD{$-$0.282}{0.074} & \DualverseMSD{122.3}{4.8} \\
\DualverseIndent\DualverseNum{3} & \DualverseNum{76.7} & \DualverseMSD{0.89441}{0.00035} & \DualverseMSD{$-$0.405}{0.071} & \DualverseMSD{132.2}{4.0} \\
\DualverseIndent\DualverseNum{4} & \DualverseNum{101.7} & \DualverseMSD{0.89421}{0.00034} & \DualverseMSD{$-$0.427}{0.071} & \DualverseMSD{134.0}{4.0} \\
\DualverseIndent\DualverseNum{16} & \DualverseNum{400.0} & \DualverseMSD{0.89415}{0.00034} & \DualverseMSD{$-$0.433}{0.071} & \DualverseMSD{134.5}{3.9} \\
\bottomrule
\end{tabularx}

%% file: sections/generated/tab_design.tex
\setlength{\DualverseColW}{1.95cm}%
\begin{tabular}{l*{5}{>{\DualverseNumT{0.00000}}C{\DualverseColW}}}
\toprule
\DualverseHeadRow
\DualverseHead[l]{Precision / codebook and scale} &
\multicolumn{3}{c}{\DualverseHead{nanochat\\\noalign{\vskip 1.6pt{\color{white}\hrule height 0.5pt}\vskip 1.6pt}
  \DualverseSubHead{\DualverseColW}{26M, 262k}\DualverseSubSep\DualverseSubHead{\DualverseColW}{26M, 1M}\DualverseSubSep\DualverseSubHead{\DualverseColW}{50M, 262k}}} &
\DualverseHead{Llama-family\\\DualverseHeadGap 31M, 262k} &
\DualverseHead{GPT-2-family\\\DualverseHeadGap 30M, 262k} \\
\midrule
\DualverseTableSection{6}{4 bits}
\DualverseIndent Uniform*, maximum-based & \DualverseNum{1.01831} & \DualverseNum{1.00783} & \DualverseNum{0.97371} & \DualverseNum{1.03994} & \DualverseNum{1.07733} \\
\DualverseIndent Lloyd--Max, error-fitted & \DualverseNum{1.01812} & \DualverseNum{1.00765} & \DualverseNum{0.97356} & \DualverseNum{1.03971} & \DualverseNum{1.07693} \\
\addlinespace[2pt]
\DualverseTableSection{6}{3 bits}
\DualverseIndent Uniform*, maximum-based & \DualverseNum{1.02004} & \DualverseNum{1.00950} & \DualverseNum{0.97499} & \DualverseNum{1.04172} & \DualverseNum{1.08039} \\
\DualverseIndent Lloyd--Max, error-fitted & \DualverseNum{1.01876} & \DualverseNum{1.00829} & \DualverseNum{0.97402} & \DualverseNum{1.04021} & \DualverseNum{1.07774} \\
\addlinespace[2pt]
\DualverseTableSection{6}{2 bits}
\DualverseIndent Uniform*, maximum-based & \DualverseNum{1.04055} & \DualverseNum{1.03016} & \DualverseNum{0.99147} & \DualverseNum{1.06998} & \DualverseNum{1.12075} \\
\DualverseIndent Uniform*, error-fitted & \DualverseNum{1.02528} & \DualverseNum{1.01509} & \DualverseNum{0.97871} & \DualverseNum{1.04720} & \DualverseNum{1.08960} \\
\DualverseIndent Uniform, maximum-based & \DualverseNum{1.02676} & \DualverseNum{1.01593} & \DualverseNum{0.98001} & \DualverseNum{1.05010} & \DualverseNum{1.13678} \\
\DualverseIndent Uniform, error-fitted & \DualverseNum{1.02170} & \DualverseNum{1.01132} & \DualverseNum{0.97611} & \DualverseNum{1.04299} & \DualverseNum{1.08240} \\
\DualverseIndent Normal quantiles, error-fitted & \DualverseNum{1.02166} & \DualverseNum{1.01128} & \DualverseNum{0.97608} & \DualverseNum{1.04296} & \DualverseNum{1.08231} \\
\DualverseIndent Lloyd--Max, RMS & \DualverseNum{1.02173} & \DualverseNum{1.01137} & \DualverseNum{0.97612} & \DualverseNum{1.04303} & \DualverseNum{1.08263} \\
\DualverseIndent Lloyd--Max, error-fitted & \DualverseNum{1.02164} & \DualverseNum{1.01127} & \DualverseNum{0.97606} & \DualverseNum{1.04295} & \DualverseNum{1.08232} \\
\addlinespace[2pt]
\DualverseTableSection{6}{1 bit}
\DualverseIndent Lloyd--Max, error-fitted & \DualverseNum{1.03963} & \DualverseNum{1.03065} & \DualverseNum{0.98935} & \DualverseNum{1.06613} & \DualverseNum{1.12362} \\
\midrule
No table & \DualverseNum{1.04855} & \DualverseNum{1.04855} & \DualverseNum{0.99403} & \DualverseNum{1.06471} & \DualverseNum{1.13947} \\
16-bit table & \DualverseNum{1.01793} & \DualverseNum{1.00746} & \DualverseNum{0.97342} & \DualverseNum{1.03956} & \DualverseNum{1.07671} \\
\bottomrule
\end{tabular}

%% file: sections/generated/tab_large_ppl.tex
\begin{tabular}{l*{2}{>{\DualverseNumT{0.0000}}C{1.65cm}}>{\DualverseNumT{+0.000}}C{1.65cm}>{\DualverseNumT{[$-$0.000, 0.000]}}C{2.8cm}}
\toprule
\DualverseHeadRow
\DualverseHead[l]{Model / dataset} & \DualverseHead{Original} & \DualverseHead{Low-bit} & \DualverseHead{Change (\%)} & \DualverseHead{95\% interval} \\
\midrule
DeepSeek / WikiText-2 & \DualverseNum{2.7564} & \DualverseNum{2.7870} & \DualverseNum{+1.110} & [\DualversePadTo{$-$0.000}{0.689}, \DualversePadTo{0.000}{1.540}] \\
DeepSeek / PTB & \DualverseNum{7.7730} & \DualverseNum{7.7794} & \DualverseNum{+0.082} & [\DualversePadTo{$-$0.000}{$-$0.145}, \DualversePadTo{0.000}{0.322}] \\
DeepSeek / C4 & \DualverseNum{7.9679} & \DualverseNum{7.9584} & \DualverseNum{$-$0.119} & [\DualversePadTo{$-$0.000}{$-$0.357}, \DualversePadTo{0.000}{0.101}] \\
Qwen / WikiText-2 & \DualverseNum{4.2270} & \DualverseNum{4.2671} & \DualverseNum{+0.949} & [\DualversePadTo{$-$0.000}{0.443}, \DualversePadTo{0.000}{1.479}] \\
Qwen / PTB & \DualverseNum{4.0633} & \DualverseNum{4.0558} & \DualverseNum{$-$0.183} & [\DualversePadTo{$-$0.000}{$-$0.656}, \DualversePadTo{0.000}{0.282}] \\
Qwen / C4 & \DualverseNum{9.2329} & \DualverseNum{9.2354} & \DualverseNum{+0.027} & [\DualversePadTo{$-$0.000}{$-$0.232}, \DualversePadTo{0.000}{0.332}] \\
\bottomrule
\end{tabular}

%% file: sections/generated/tab_large_downstream.tex
\begin{tabular}{l*{2}{>{\DualverseNumT{00.00}}C{1.5cm}}>{\DualverseNumT{+0.00}}C{1.65cm}>{\DualverseNumT{[$-$0.00, +0.00]}}C{2.6cm}}
\toprule
\DualverseHeadRow
\DualverseHead[l]{Model / task} & \DualverseHead{Original} & \DualverseHead{Low-bit} & \DualverseHead{Difference} & \DualverseHead{95\% interval} \\
\midrule
DeepSeek / HellaSwag & \DualverseNum{82.48} & \DualverseNum{82.44} & \DualverseNum{$-$0.04} & [\DualversePadTo{$-$0.00}{$-$0.38}, \DualversePadTo{+0.00}{+0.29}] \\
DeepSeek / ARC-Challenge & \DualverseNum{61.60} & \DualverseNum{60.58} & \DualverseNum{$-$1.02} & [\DualversePadTo{$-$0.00}{$-$2.65}, \DualversePadTo{+0.00}{+0.60}] \\
DeepSeek / WinoGrande & \DualverseNum{64.48} & \DualverseNum{62.75} & \DualverseNum{$-$1.74} & [\DualversePadTo{$-$0.00}{$-$4.18}, \DualversePadTo{+0.00}{+0.71}] \\
Qwen / HellaSwag & \DualverseNum{87.72} & \DualverseNum{87.57} & \DualverseNum{$-$0.15} & [\DualversePadTo{$-$0.00}{$-$0.39}, \DualversePadTo{+0.00}{+0.09}] \\
Qwen / ARC-Challenge & \DualverseNum{62.88} & \DualverseNum{63.65} & \DualverseNum{+0.77} & [\DualversePadTo{$-$0.00}{$-$0.68}, \DualversePadTo{+0.00}{+2.22}] \\
Qwen / WinoGrande & \DualverseNum{72.77} & \DualverseNum{71.35} & \DualverseNum{$-$1.42} & [\DualversePadTo{$-$0.00}{$-$3.24}, \DualversePadTo{+0.00}{+0.39}] \\
\bottomrule
\end{tabular}

%% file: sections/generated/tab_timing.tex
\begin{tabular}{l*{4}{>{\DualversePMT{$-$00.0}{0.0}}C{1.85cm}}}
\toprule
\DualverseHeadRow
\DualverseHead[l]{Training sequence} & \DualverseHead{1 bit} & \DualverseHead{2 bits} & \DualverseHead{3 bits} & \DualverseHead{4 bits} \\
\midrule
Compress $\to$ pre-train & \DualverseMSD{77.0}{4.6} & \DualverseMSD{87.0}{9.4} & \DualverseMSD{92.2}{9.3} & \DualverseMSD{96.8}{0.3} \\
Pre-train $\to$ compress $\to$ fine-tune & \DualverseMSD{65.3}{0.9} & \DualverseMSD{92.0}{0.3} & \DualverseMSD{98.0}{0.0} & \DualverseMSD{99.5}{0.0} \\
Pre-train $\to$ fine-tune $\to$ compress & \DualverseMSD{$-$5.7}{6.3} & \DualverseMSD{86.5}{0.9} & \DualverseMSD{97.4}{0.2} & \DualverseMSD{99.4}{0.1} \\
Pre-train $\to$ compress & \DualverseMSD{$-$5.7}{4.6} & \DualverseMSD{86.5}{0.4} & \DualverseMSD{97.4}{0.1} & \DualverseMSD{99.4}{0.0} \\
\bottomrule
\end{tabular}

%% file: sections/generated/tab_rows_finetune.tex
\setlength{\DualverseColW}{3.1cm}%
\begin{tabular}{>{\DualverseNumT{00}}l*{2}{>{\DualversePMT{000.0}{0.0}}C{\DualverseColW}}>{\DualverseNumT{000.0}}C{2.0cm}}
\toprule
\DualverseHeadRow
\DualverseHead[l]{Precision\\(bits)} &
\multicolumn{2}{c}{\DualverseHead{Improvement relative to original (\%)\\\noalign{\vskip 1.6pt{\color{white}\hrule height 0.5pt}\vskip 1.6pt}
  \DualverseSubHead{\DualverseColW}{Fine-tune then\\compress}\DualverseSubSep\DualverseSubHead{\DualverseColW}{Compress then\\fine-tune}}} &
\DualverseHead{Memory\\(\%)} \\
\midrule
\DualverseTableSection{4}{nanochat 26M}
\DualverseIndent\DualverseNum{1} & \DualverseMSD{63.0}{2.1} & \DualverseMSD{97.4}{0.9} & \DualverseNum{29.2} \\
\DualverseIndent\DualverseNum{2} & \DualverseMSD{126.0}{2.8} & \DualverseMSD{128.7}{3.0} & \DualverseNum{54.2} \\
\DualverseIndent\DualverseNum{3} & \DualverseMSD{135.5}{3.4} & \DualverseMSD{135.8}{3.5} & \DualverseNum{79.2} \\
\DualverseIndent\DualverseNum{4} & \DualverseMSD{137.5}{3.6} & \DualverseMSD{137.6}{3.6} & \DualverseNum{104.2} \\
\addlinespace[2pt]
\DualverseTableSection{4}{Llama-family 31M}
\DualverseIndent\DualverseNum{1} & \DualverseMSD{44.2}{4.4} & \DualverseMSD{110.8}{2.5} & \DualverseNum{28.1} \\
\DualverseIndent\DualverseNum{2} & \DualverseMSD{131.1}{2.0} & \DualverseMSD{136.3}{1.8} & \DualverseNum{53.1} \\
\DualverseIndent\DualverseNum{3} & \DualverseMSD{141.2}{1.9} & \DualverseMSD{141.8}{1.8} & \DualverseNum{78.1} \\
\DualverseIndent\DualverseNum{4} & \DualverseMSD{143.0}{1.8} & \DualverseMSD{143.1}{1.7} & \DualverseNum{103.1} \\
\addlinespace[2pt]
\DualverseTableSection{4}{Llama-family 322M}
\DualverseIndent\DualverseNum{1} & \DualverseMSD{45.2}{9.3} & \DualverseMSD{106.0}{5.0} & \DualverseNum{26.7} \\
\DualverseIndent\DualverseNum{2} & \DualverseMSD{141.3}{3.5} & \DualverseMSD{144.7}{3.5} & \DualverseNum{51.7} \\
\DualverseIndent\DualverseNum{3} & \DualverseMSD{150.6}{3.2} & \DualverseMSD{150.9}{3.3} & \DualverseNum{76.7} \\
\DualverseIndent\DualverseNum{4} & \DualverseMSD{152.3}{3.3} & \DualverseMSD{152.3}{3.3} & \DualverseNum{101.7} \\
\bottomrule
\end{tabular}

%% file: sections/generated/tab_hashing.tex
\begin{tabular}{l*{3}{>{\DualversePMT{000.00}{00.00}}C{3.05cm}}}
\toprule
\DualverseHeadRow
\DualverseHead[l]{Model} & \DualverseHead{Conventional} & \DualverseHead{Same-column} & \DualverseHead{Fully random} \\
\midrule
\DualverseTableSection{4}{16 bits}
\DualverseIndent nanochat 26M & \DualverseMSD{100.00}{0.00} & \DualverseMSD{104.84}{1.14} & \DualverseMSD{102.14}{4.33} \\
\DualverseIndent GPT-2-family 30M & \DualverseMSD{100.00}{0.00} & \DualverseMSD{98.46}{2.95} & \DualverseMSD{99.92}{3.36} \\
\DualverseIndent Llama-family 31M & \DualverseMSD{100.00}{0.00} & \DualverseMSD{112.03}{4.53} & \DualverseMSD{107.97}{10.99} \\
\addlinespace[2pt]
\DualverseTableSection{4}{2 bits}
\DualverseIndent nanochat 26M & \DualverseMSD{91.71}{1.23} & \DualverseMSD{93.61}{2.36} & \DualverseMSD{93.95}{3.04} \\
\DualverseIndent GPT-2-family 30M & \DualverseMSD{93.69}{2.20} & \DualverseMSD{99.57}{5.48} & \DualverseMSD{100.16}{6.92} \\
\DualverseIndent Llama-family 31M & \DualverseMSD{87.03}{9.41} & \DualverseMSD{100.28}{10.54} & \DualverseMSD{98.39}{1.81} \\
\bottomrule
\end{tabular}

%% file: paper.bbl
\begin{thebibliography}{35}
\providecommand{\natexlab}[1]{#1}
\providecommand{\url}[1]{\texttt{#1}}
\expandafter\ifx\csname urlstyle\endcsname\relax
  \providecommand{\doi}[1]{doi: #1}\else
  \providecommand{\doi}{doi: \begingroup \urlstyle{rm}\Url}\fi

\bibitem[Huang et~al.(2025)Huang, Zhu, Wu, Zeng, Wang, Min, and
  Zhou]{huang2025overtokenized}
Hongzhi Huang, Defa Zhu, Banggu Wu, Yutao Zeng, Ya~Wang, Qiyang Min, and Xun
  Zhou.
\newblock Over-tokenized transformer: Vocabulary is generally worth scaling.
\newblock In \emph{Proceedings of the 42nd International Conference on Machine
  Learning}, volume 267 of \emph{Proceedings of Machine Learning Research},
  pages 26261--26282, 2025.

\bibitem[Cheng et~al.(2026)Cheng, Zeng, Dai, Chen, Wang, Xie, Huang, Yu, Hao,
  Li, Zhang, Zhang, Zhao, and Liang]{cheng2026dse}
Xin Cheng, Wangding Zeng, Damai Dai, Qinyu Chen, Bingxuan Wang, Zhenda Xie,
  Kezhao Huang, Xingkai Yu, Zhewen Hao, Yukun Li, Han Zhang, Huishuai Zhang,
  Dongyan Zhao, and Wenfeng Liang.
\newblock Conditional memory via scalable lookup: A new axis of sparsity for
  large language models.
\newblock In \emph{Proceedings of the 64th Annual Meeting of the Association
  for Computational Linguistics (Volume 1: Long Papers)}, pages 4968--4990,
  2026.
\newblock URL \url{https://aclanthology.org/2026.acl-long.226/}.

\bibitem[{DeepSeek-AI}(2026)]{deepseek2026v41flash}
{DeepSeek-AI}.
\newblock {DeepSeek-V4.1-Flash}: Model card.
\newblock \url{https://huggingface.co/deepseek-ai/DeepSeek-V4.1-Flash}, 2026.
\newblock Accessed October 19, 2026.

\bibitem[Lloyd(1982)]{lloyd1982least}
Stuart~P. Lloyd.
\newblock Least squares quantization in {PCM}.
\newblock \emph{IEEE Transactions on Information Theory}, 28\penalty0
  (2):\penalty0 129--137, 1982.

\bibitem[Max(1960)]{max1960quantizing}
Joel Max.
\newblock Quantizing for minimum distortion.
\newblock \emph{IRE Transactions on Information Theory}, 6\penalty0
  (1):\penalty0 7--12, 1960.

\bibitem[Karpathy(2025)]{karpathy2025nanochat}
Andrej Karpathy.
\newblock nanochat.
\newblock \url{https://github.com/karpathy/nanochat}, 2025.

\bibitem[Karpathy(2026)]{karpathy2026autoresearch}
Andrej Karpathy.
\newblock autoresearch.
\newblock \url{https://github.com/karpathy/autoresearch}, commit 228791f, 2026.

\bibitem[Touvron et~al.(2023)Touvron, Lavril, Izacard, Martinet, Lachaux,
  Lacroix, Rozi{\`e}re, Goyal, Hambro, Azhar, et~al.]{touvron2023llama}
Hugo Touvron, Thibaut Lavril, Gautier Izacard, Xavier Martinet, Marie-Anne
  Lachaux, Timoth{\'e}e Lacroix, Baptiste Rozi{\`e}re, Naman Goyal, Eric
  Hambro, Faisal Azhar, et~al.
\newblock {LLaMA}: Open and efficient foundation language models.
\newblock \emph{arXiv preprint arXiv:2302.13971}, 2023.

\bibitem[Radford et~al.(2019)Radford, Wu, Child, Luan, Amodei, and
  Sutskever]{radford2019gpt2}
Alec Radford, Jeffrey Wu, Rewon Child, David Luan, Dario Amodei, and Ilya
  Sutskever.
\newblock Language models are unsupervised multitask learners.
\newblock Technical report, OpenAI, 2019.

\bibitem[Diao et~al.(2025)Diao, Yang, Fu, Dong, Su, Kliegl, Chen, Belcak,
  Suhara, Yin, Patwary, Lin, Kautz, and Molchanov]{diao2025climb}
Shizhe Diao, Yu~Yang, Yonggan Fu, Xin Dong, Dan Su, Markus Kliegl, Zijia Chen,
  Peter Belcak, Yoshi Suhara, Hongxu Yin, Mostofa Patwary, Yingyan Lin, Jan
  Kautz, and Pavlo Molchanov.
\newblock {Nemotron-CLIMB}: Clustering-based iterative data mixture
  bootstrapping for language model pre-training.
\newblock In \emph{Advances in Neural Information Processing Systems},
  volume~38, 2025.

\bibitem[{Qwen Team}(2026)]{qwen2026flashnext}
{Qwen Team}.
\newblock {Qwen3.8-Flash-Next}: Model card.
\newblock \url{https://huggingface.co/Qwen/Qwen3.8-Flash-Next}, 2026.
\newblock Accessed October 20, 2026.

\bibitem[Merity et~al.(2017)Merity, Xiong, Bradbury, and
  Socher]{merity2017pointer}
Stephen Merity, Caiming Xiong, James Bradbury, and Richard Socher.
\newblock Pointer sentinel mixture models.
\newblock In \emph{International Conference on Learning Representations}, 2017.

\bibitem[Marcus et~al.(1993)Marcus, Santorini, and
  Marcinkiewicz]{marcus1993building}
Mitchell~P. Marcus, Beatrice Santorini, and Mary~Ann Marcinkiewicz.
\newblock Building a large annotated corpus of english: The {Penn Treebank}.
\newblock \emph{Computational Linguistics}, 19\penalty0 (2):\penalty0 313--330,
  1993.

\bibitem[Raffel et~al.(2020)Raffel, Shazeer, Roberts, Lee, Narang, Matena,
  Zhou, Li, and Liu]{raffel2020exploring}
Colin Raffel, Noam Shazeer, Adam Roberts, Katherine Lee, Sharan Narang, Michael
  Matena, Yanqi Zhou, Wei Li, and Peter~J. Liu.
\newblock Exploring the limits of transfer learning with a unified text-to-text
  transformer.
\newblock \emph{Journal of Machine Learning Research}, 21\penalty0
  (140):\penalty0 1--67, 2020.

\bibitem[Zellers et~al.(2019)Zellers, Holtzman, Bisk, Farhadi, and
  Choi]{zellers2019hellaswag}
Rowan Zellers, Ari Holtzman, Yonatan Bisk, Ali Farhadi, and Yejin Choi.
\newblock {HellaSwag}: Can a machine really finish your sentence?
\newblock In \emph{Proceedings of the 57th Annual Meeting of the Association
  for Computational Linguistics}, pages 4791--4800, 2019.

\bibitem[Clark et~al.(2018)Clark, Cowhey, Etzioni, Khot, Sabharwal, Schoenick,
  and Tafjord]{clark2018arc}
Peter Clark, Isaac Cowhey, Oren Etzioni, Tushar Khot, Ashish Sabharwal, Carissa
  Schoenick, and Oyvind Tafjord.
\newblock Think you have solved question answering? try {ARC}, the {AI2}
  reasoning challenge, 2018.

\bibitem[Sakaguchi et~al.(2020)Sakaguchi, Le~Bras, Bhagavatula, and
  Choi]{sakaguchi2020winogrande}
Keisuke Sakaguchi, Ronan Le~Bras, Chandra Bhagavatula, and Yejin Choi.
\newblock {WinoGrande}: An adversarial winograd schema challenge at scale.
\newblock In \emph{Proceedings of the AAAI Conference on Artificial
  Intelligence}, volume~34, pages 8732--8740, 2020.

\bibitem[Bengio et~al.(2003)Bengio, Ducharme, Vincent, and
  Jauvin]{bengio2003neural}
Yoshua Bengio, R{\'e}jean Ducharme, Pascal Vincent, and Christian Jauvin.
\newblock A neural probabilistic language model.
\newblock \emph{Journal of Machine Learning Research}, 3:\penalty0 1137--1155,
  2003.

\bibitem[Chen et~al.(2015)Chen, Wilson, Tyree, Weinberger, and
  Chen]{chen2015hashing}
Wenlin Chen, James Wilson, Stephen Tyree, Kilian Weinberger, and Yixin Chen.
\newblock Compressing neural networks with the hashing trick.
\newblock In \emph{Proceedings of the 32nd International Conference on Machine
  Learning}, volume~37 of \emph{Proceedings of Machine Learning Research},
  2015.

\bibitem[Svenstrup et~al.(2017)Svenstrup, Hansen, and
  Winther]{svenstrup2017hash}
Dan~Tito Svenstrup, Jonas Hansen, and Ole Winther.
\newblock Hash embeddings for efficient word representations.
\newblock In \emph{Advances in Neural Information Processing Systems},
  volume~30, 2017.

\bibitem[Roy et~al.(2022)Roy, Anil, Lai, Lee, Zhao, Zhang, Wang, Zhang, Wu,
  Swavely, et~al.]{roy2022ngrammer}
Aurko Roy, Rohan Anil, Guangda Lai, Benjamin Lee, Jeffrey Zhao, Shuyuan Zhang,
  Shibo Wang, Ye~Zhang, Shen Wu, Rigel Swavely, et~al.
\newblock {N-Grammer}: Augmenting transformers with latent n-grams.
\newblock \emph{arXiv preprint arXiv:2207.06366}, 2022.

\bibitem[Pagnoni et~al.(2025)Pagnoni, Pasunuru, Rodriguez, Nguyen, Muller, Li,
  Zhou, Yu, Weston, Zettlemoyer, Ghosh, Lewis, Holtzman, and
  Iyer]{pagnoni2025blt}
Artidoro Pagnoni, Ram Pasunuru, Pedro Rodriguez, John Nguyen, Benjamin Muller,
  Margaret Li, Chunting Zhou, Lili Yu, Jason Weston, Luke Zettlemoyer, Gargi
  Ghosh, Mike Lewis, Ari Holtzman, and Srinivasan Iyer.
\newblock Byte latent transformer: Patches scale better than tokens.
\newblock In \emph{Proceedings of the 63rd Annual Meeting of the Association
  for Computational Linguistics}, pages 9238--9258, 2025.

\bibitem[Zhou et~al.(2026)Zhou, Gu, Iacovides, Qiu, Zhao, and
  Mandic]{zhou2026tngram}
Wuyang Zhou, Yuxuan Gu, Giorgos Iacovides, Yuning Qiu, Qibin Zhao, and Danilo
  Mandic.
\newblock Tensorizing engram: Sharing latents across n-gram embeddings is
  beneficial in {LLMs}.
\newblock \emph{arXiv preprint arXiv:2606.08347}, 2026.

\bibitem[Yang et~al.(2026)Yang, Zhou, Miao, and Wu]{yang2026factorengram}
Bowen Yang, Jingbo Zhou, Qinghong Miao, and Hua Wu.
\newblock {FactorEngram}: Factorized n-gram memory with basis-level gating for
  language models.
\newblock \emph{arXiv preprint arXiv:2609.35578}, 2026.

\bibitem[Chen et~al.(2026)Chen, Xie, Gao, Xin, Xiao, Liu, Luo, Luo, Ye, Liu,
  Zhao, Tao, and Dai]{chen2026xgram}
Yilong Chen, Yanxi Xie, Zitian Gao, He~Xin, Yihao Xiao, Jason~Klein Liu,
  Haoming Luo, Yifan Luo, Zhengmao Ye, Tingwen Liu, Xin Zhao, Ran Tao, and
  Bryan Dai.
\newblock Beyond n-gram: Data-aware {X-GRAM} extraction for efficient embedding
  parameter scaling.
\newblock \emph{arXiv preprint arXiv:2604.21724}, 2026.

\bibitem[Yu et~al.(2025)Yu, Cohen, Ghazi, Huang, Kamath, Kumar, Liu, and
  Zhang]{yu2025scone}
Da~Yu, Edith Cohen, Badih Ghazi, Yangsibo Huang, Pritish Kamath, Ravi Kumar,
  Daogao Liu, and Chiyuan Zhang.
\newblock Scaling embedding layers in language models.
\newblock In \emph{Advances in Neural Information Processing Systems},
  volume~38, pages 31969--31995, 2025.

\bibitem[Lample et~al.(2019)Lample, Sablayrolles, Ranzato, Denoyer, and
  J{\'e}gou]{lample2019product}
Guillaume Lample, Alexandre Sablayrolles, Marc'Aurelio Ranzato, Ludovic
  Denoyer, and Herv{\'e} J{\'e}gou.
\newblock Large memory layers with product keys.
\newblock In \emph{Advances in Neural Information Processing Systems},
  volume~32, 2019.

\bibitem[Berges et~al.(2025)Berges, O{\u{g}}uz, Haziza, Yih, Zettlemoyer, and
  Ghosh]{berges2025memory}
Vincent-Pierre Berges, Barlas O{\u{g}}uz, Daniel Haziza, Wen-tau Yih, Luke
  Zettlemoyer, and Gargi Ghosh.
\newblock Memory layers at scale.
\newblock In \emph{Proceedings of the 42nd International Conference on Machine
  Learning}, volume 267 of \emph{Proceedings of Machine Learning Research},
  pages 3831--3842, 2025.

\bibitem[Khandelwal et~al.(2020)Khandelwal, Levy, Jurafsky, Zettlemoyer, and
  Lewis]{khandelwal2020knnlm}
Urvashi Khandelwal, Omer Levy, Dan Jurafsky, Luke Zettlemoyer, and Mike Lewis.
\newblock Generalization through memorization: Nearest neighbor language
  models.
\newblock In \emph{International Conference on Learning Representations}, 2020.

\bibitem[Borgeaud et~al.(2022)Borgeaud, Mensch, Hoffmann, Cai, Rutherford,
  Millican, et~al.]{borgeaud2022retro}
Sebastian Borgeaud, Arthur Mensch, Jordan Hoffmann, Trevor Cai, Eliza
  Rutherford, Katie Millican, et~al.
\newblock Improving language models by retrieving from trillions of tokens.
\newblock In \emph{Proceedings of the 39th International Conference on Machine
  Learning}, volume 162 of \emph{Proceedings of Machine Learning Research},
  2022.

\bibitem[Guan et~al.(2019)Guan, Malevich, Yang, Park, and
  Yuen]{guan2019embedding}
Hui Guan, Andrey Malevich, Jiyan Yang, Jongsoo Park, and Hector Yuen.
\newblock Post-training 4-bit quantization on embedding tables.
\newblock In \emph{MLSys Workshop at Advances in Neural Information Processing
  Systems}, 2019.

\bibitem[Shi et~al.(2020)Shi, Mudigere, Naumov, and Yang]{shi2020compositional}
Hao-Jun~Michael Shi, Dheevatsa Mudigere, Maxim Naumov, and Jiyan Yang.
\newblock Compositional embeddings using complementary partitions for
  memory-efficient recommendation systems.
\newblock In \emph{Proceedings of the 26th ACM SIGKDD International Conference
  on Knowledge Discovery and Data Mining}, pages 165--175, 2020.

\bibitem[Ginart et~al.(2021)Ginart, Naumov, Mudigere, Yang, and
  Zou]{ginart2021mixed}
Antonio~A. Ginart, Maxim Naumov, Dheevatsa Mudigere, Jiyan Yang, and James Zou.
\newblock Mixed dimension embeddings with application to memory-efficient
  recommendation systems.
\newblock In \emph{IEEE International Symposium on Information Theory}, pages
  2786--2791, 2021.

\bibitem[Shu and Nakayama(2018)]{shu2018compositional}
Raphael Shu and Hideki Nakayama.
\newblock Compressing word embeddings via deep compositional code learning.
\newblock In \emph{International Conference on Learning Representations}, 2018.

\bibitem[Ben~Allal et~al.(2025)Ben~Allal, Lozhkov, Bakouch,
  Mart{\'\i}n~Bl{\'a}zquez, Penedo, Tunstall, et~al.]{allal2025smollm2}
Loubna Ben~Allal, Anton Lozhkov, Elie Bakouch, Gabriel
  Mart{\'\i}n~Bl{\'a}zquez, Guilherme Penedo, Lewis Tunstall, et~al.
\newblock {SmolLM2}: When smol goes big -- data-centric training of a fully
  open small language model.
\newblock In \emph{Second Conference on Language Modeling}, 2025.
\newblock URL \url{https://openreview.net/forum?id=3JiCl2A14H}.

\end{thebibliography}
